\documentclass[runningheads]{llncs}

\usepackage{xcolor}

\usepackage[colorlinks = true,
            linkcolor = blue,
           urlcolor  = blue,
           citecolor = blue,
            anchorcolor = blue,
           bookmarks=false,
           hypertexnames=true]{hyperref}
\usepackage{subfigure}
\usepackage{enumitem}
\usepackage[ruled,vlined,noend,linesnumbered]{algorithm2e}
\usepackage{amsmath}
\usepackage{tikz}
\usepackage[labelfont=bf]{caption}
\usepackage{multicol}
\usepackage{multirow}
\usepackage{array}
\usepackage{booktabs}
\newcolumntype{L}[1]{>{\raggedright\let\newline\\\arraybackslash\hspace{0pt}}m{#1}}
\newcolumntype{C}[1]{>{\centering\let\newline\\\arraybackslash\hspace{0pt}}m{#1}}
\newcolumntype{R}[1]{>{\raggedleft\let\newline\\\arraybackslash\hspace{0pt}}m{#1}}

\newcommand{\prob}{\ensuremath{\mathsf{Pr}}}
\newcommand{\satisfying}[1]{\ensuremath{R_{#1}}} %
\newcommand{\tool}{\ensuremath{\mathsf{Manthan}}}

\newcommand{\bfss}{BFSS}
\newcommand{\ctosyn}{C2Syn}
\newcommand{\cadet}{CADET}
\newcommand{\bafsyn}{BaFSyn}
\newcommand{\unsat}{{UNSAT}}
\newcommand{\unsatcore}{{UnsatCore}}
\newcommand{\sat}{{SAT}}
\newcommand{\maxsat}{{MaxSAT}}
\newcommand{\caqe}{{CAQE}}
\newcommand{\depqbf}{DepQBF}
\newcommand{\totalorder}{\textit{TotalOrder}}

\newcommand{\sampler}{BiasGen}
\newcommand{\adasampler}{\ensuremath{\mathsf{AdaBiasGen}}}
\newcommand{\phaseone}{\ensuremath{\mathsf{Preprocess}}}
\newcommand{\phasetwo}{\ensuremath{\mathsf{LearnSkF}}}
\newcommand{\phasethree}{\ensuremath{\mathsf{Refine}}}

\newcommand{\candidateskf}{\ensuremath{\mathsf{CandidateSkF}}}
\newcommand{\bias}{\ensuremath{\mathsf{Bias}}}
\newcommand{\preprocess}{\ensuremath{\mathsf{Preprocess}}}
\newcommand{\findorder}{\ensuremath{\mathsf{FindOrder}}}
\newcommand{\checksat}{\ensuremath{\mathsf{CheckSat}}}
\newcommand{\refine}{\ensuremath{\mathsf{RefineSkF}}}
\newcommand{\substitute}{\ensuremath{\mathsf{Substitute}}}
\newcommand{\getsamples}{\ensuremath{\mathsf{GetSamples}}}

\newcommand{\createdt}{\ensuremath{\mathsf{CreateDecisionTree}}}

\newcommand{\Path}{\ensuremath{\mathsf{Path}}}

\newcommand{\Label}{\ensuremath{\mathsf{Label}}}

\newcommand{\checksub}{\ensuremath{\mathsf{CheckSubstitute}}}
\newcommand{\selfsub}{\ensuremath{\mathsf{DoSelfSubstitution}}}
\newcommand{\findcore}{\ensuremath{\mathsf{FindCore}}}
\newcommand{\maxsatlist}{\ensuremath{\mathsf{MaxSATList}}}

\usepackage[draft]{todonotes}   % notes showed

\begin{document}
\title{Manthan: A Data Driven Approach for Boolean Function Synthesis\thanks{The open source tool is available at \url{https://github.com/meelgroup/manthan}}}
%\title{Manthan: A Data Driven Approach for Boolean Function Synthesis\thanks{The supplementary material consisting of Appendix and detailed experimental data is available at \url{https://www.cse.iitk.ac.in/users/pgolia/manthan/}; Password: ``manthan" (without quotes). In line with prior work, the tool will be released as open-source via Github at \url{https://github.com/meelgroup/} post-publication.}}

%\author{Anonymous Authors}
\institute{}
\author{Priyanka Golia\inst{1,2} \and  Subhajit Roy\inst{1} \and Kuldeep S. Meel\inst{2}}
\authorrunning{Golia, Roy, and Meel}
\institute{ Computer Science and Engineering, Indian Institute of Technology Kanpur, India \email{{pgolia,subhajit}@cse.iitk.ac.in} \and School of Computing, National University of Singapore, Singapore\\
	\email{meel@comp.nus.edu.sg}}

\maketitle              % typeset the header of the contribution
\begin{abstract}
    Boolean functional synthesis is a fundamental problem in computer science with wide-ranging applications and has witnessed a surge of interest resulting in progressively improved techniques over the past decade. Despite intense algorithmic development, a large number of problems remain beyond the reach of the state of the art techniques.

    Motivated by the progress in machine learning, we propose {\tool}, a novel data-driven approach to Boolean functional synthesis. {\tool} views functional synthesis as a classification problem, relying on advances in constrained sampling for data generation, and advances in automated reasoning for a novel proof-guided refinement and provable verification. On an extensive and rigorous evaluation over 609 benchmarks, we demonstrate that {\tool} significantly improves upon the current state of the art, solving 356 benchmarks in comparison to 280, which is the most  solved by a state of the art technique; thereby, we demonstrate an increase of 76 benchmarks over the current state of the art. Furthermore, {\tool} solves 60 benchmarks that none of the current state of the art techniques could solve. The significant performance improvements, along with our detailed analysis, highlights several interesting avenues of future work at the intersection of machine learning, constrained sampling, and automated reasoning. 
\end{abstract}

\section{Introduction} 
Given an existentially quantified Boolean formula $\exists Y F(X,Y)$ over the set of variables $X$ and $Y$, the problem of Boolean functional synthesis is to compute a vector of Boolean functions, denoted by $\Psi(X) = \langle\psi_{1}(X), \psi_{2}(X), \ldots ,\psi_{|Y|}(X)\rangle$, and referred to as Skolem function vector, such that $\exists Y F(X,Y) \equiv F(X, \Psi(X))$. In the context of applications, the sets $X$ and $Y$ are viewed as inputs and outputs, and the formula $F(X,Y)$ is viewed as a functional specification capturing the relationship between $X$ and $Y$, while the Skolem function vector $\Psi(X)$ allows one to determine the value of $Y$ for the given $X$ by evaluating $\Psi$. The study of Boolean functional synthesis traces back to Boole~\cite{boole1847}, and over the decades, the problem has found applications in a wide variety of domains such as certified QBF solving~\cite{BJ11,BJ12,NPLSB12,RTRS18}, automated program repair~\cite{JMF14}, program synthesis~\cite{SGF13}, and cryptography~\cite{MM20}.
%See references in input-output paper

Theoretical investigations have demonstrated that there exist instances where %$\mathsf{Co-NP}^{\mathsf{NP}}$-hardness of%
 Boolean functional synthesis takes super-polynomial time. On the other hand,  practical applicability has necessitated the  development of algorithms with progressively impressive scaling.
The algorithmic progress for Boolean functional synthesis has been driven by a diverse set of techniques: (i) the usage of incremental determinization employing the several heuristics in state-of-the-art Conflict Driven Clause Learning (CDCL) solvers~\cite{RTRS18}, (ii) usage of decomposition techniques employing the progress in knowledge compilation~\cite{ACJS17,FTV16,JSCTA15,TV17}, and (iii) Counter-Example Guided Abstraction Refinement (CEGAR)-based techniques relying on usage of SAT solvers as black boxes~\cite{AACKRS19,ACJS17,ACGKS18,JSCTA15}. While the state of the art techniques are capable of handling problems of complexity beyond the capability of tools a decade ago, the design of scalable algorithms capable of handling industrial problems remains the holy grail.

In this work, we take a step towards the above goal by proposing a novel approach, called {\tool}, at the intersection of machine learning, constrained sampling, and automated reasoning. Motivated by the unprecedented advances in machine learning, we view the problem of functional synthesis through the lens of multi-class classification aided by the generation of the data via constrained sampling and employ automated reasoning to certify and refine the learned functions. To this end, the architecture of {\tool} comprises of the following three novel techniques:
\begin{description}[topsep=0pt]
	\item[Data Generation] The state of the art machine learning techniques use training data represented as a set of samples where each sample consists of valuations to features and the corresponding label. In our context, we treat $X$ as the features and $Y$ as labels. Unlike the standard setup of machine learning wherein for each assignment to $X$, there is a unique label, i.e. assignment to $Y$, the relationship between $X$ and $Y$ is captured by a relation and not necessarily a function. To this end, we design a weighted sampling strategy to generate a {\em representative} data set that can be fitted using a {\em compactly sized} classifier. The weighted sampling strategy, implemented using state of the constrained sampler, seeks to uniformly sample input variables ($X$) while biasing the valuations of output variables towards a particular value. 
	
	\item[Dependency-Driven Classifier for Candidates] Given training data viewed as a valuation of {\em features} (X) and their corresponding labels (Y),  a natural approach from machine learning perspective would be to perform multi-class classification to obtain $Y = h (X)$, where $h$ is a symbolic representation of the learned classifier. Such an approach, however, can not ensure that $h$ can be expressed as a vector of Boolean functions. To this end, we design a dependency aware classifier to construct a vector of decision trees corresponding to each $Y_i$, wherein each decision tree is expressed as a Boolean function. 
	
	\item[Proof-Guided Refinement] Since machine learning techniques often produce good but inexact approximations, we augment our method with automated reasoning techniques to verify the correctness of decision tree-based candidate Skolem functions. To this end, we perform a counterexample driven refinement approach for candidate Skolem functions. 
	
	To fully utilize the impressive test accuracy attained by machine learning models, we design a {\em proof-guided refinement} approach that seeks to identify and apply {\em minor} repairs to the candidate functions, in an iterative manner, until we converge to a provably correct Skolem function vector. In a departure from prior approaches utilizing the Shannon expansion and self-substitution,  we first use a {\maxsat} solver to determine potential repair candidates, and employ unsatisfiability cores obtained from the infeasibility proofs capturing the reason for current candidate functions to meet the specification, to construct a {\em good repair}.            
\end{description}

Finally, We perform an extensive evaluation over a diverse set of benchmarks with state-of-the-art tools, viz. {\ctosyn}\cite{AACKRS19}, {\bfss}\cite{ACGKS18}, and {\cadet}\cite{R19}. Of 609 benchmarks, {\tool} is able to solve $356$ benchmarks  while {\ctosyn}, {\bfss}, and {\cadet} solve $206$, $247$, and $280$ benchmarks respectively. Significantly, {\tool} can solve $60$ benchmarks beyond the reach of all the other existing tools extending the reach of functional synthesis tools. We then perform an extensive empirical evaluation to understand the impact of different design choices on the performance of {\tool}. Our study reveals several surprising observations arising from the inter-play of machine learning and automated reasoning.

{\tool} owes its runtime performance to recent advances in machine learning, constrained sampling, and automated reasoning. Encouraged by {\tool}'s scalability, we will seek to extend the above approach to related problem domains such as automated program synthesis, program repair, and reactive synthesis. 
%\par

The rest of the paper is organized as follows: We first introduce notations and preliminaries in Section~\ref{preliminaries}. We then discuss the related work in Section~\ref{sec:relatedwork}. In Section~\ref{sec:overview} we present an overview of {\tool} and give an algorithmic description in Section ~\ref{sec:algorithm}. We then describe the experimental methodology and discuss results in Section~\ref{sec:experiments}. Finally, we conclude in Section~\ref{sec:conclusion}.

\section{Notations and Preliminaries}
\label{preliminaries}

%We use usual boolean notations  $\lnot$, $\land$, $\lor$ to denote negation, conjunction and disjunction. We use symbol $\equiv$ to denote logical equivalence between two boolean formulas. 
We use lower case letters (with subscripts) to denote propositional variables and upper case letters to denote a subset of variables. The formula $ \exists Y F(X,Y)$ is existentially quantified in $Y$, where $X=\{x_1,\cdots,x_n\}$ and $Y=\{y_1,\cdots,y_m\}$. For notational clarity, we use $F$ to refer to $F(X,Y)$ when clear from the context. We denote $Vars(F)$ as the set of variables appearing in $F(X,Y)$. A literal is a boolean variable or its negation. We often abbreviate universally (resp. existentially) quantified variables as universal (resp. existential) variables.

A \textit{satisfying assignment} of a formula $F(X,Y)$ is a mapping $\sigma: Vars(F) \rightarrow \{0,1\}$, on which the formula evaluates to True.  For $V \subseteq Vars(F)$, $\sigma[V]$ represents the truth values of variables in $V$ in a satisfying assignment $\sigma$ of $F$. We denote the set of all witnesses of $F$ as $\satisfying{F}$.
%A formula $F(X,Y)$ is \textit{satisfiable}({\sat}), if there exists a satisfying assignment of $F(X,Y)$; else it is \textit{unsatisfiable}({\unsat}). 
For a formula in conjunctive normal form, the \textit{unsatisfiable core}({\unsatcore}) is a subset of clauses of the formula for which no satisfying assignment exists.

We use $F(X,Y)|_{y_i=b}$ to denote \textit{substitutions}: a formula obtained after substituting every occurrence of $y_i$ in $F(X,Y)$ by $b$, where $b$ can be a constant ($0$ or $1$) or a formula. The operator \textit{ite(condition,exp1,exp2)} is used to represent the if-else case: if the \textit{condition} is true, then it returns \textit{exp1}, else it returns \textit{exp2}.

			%\begin{multline}
				%\label{eqn:errorformula}
				%E(X,Y,Y') = F(X,Y) \land \lnot F(X,Y') \land (Y'\leftrightarrow \Psi)
				%\\
				%where \; Y'=\{y'_1,\cdots,y'_{|Y|}\} \;  \; s.t. \; Y' \cap Y = \phi
			%\end{multline}

\label{pro:unates}A variable $y_i$ is considered as a \textit{positive unate} if and only if $F(X,Y)|_{y_i=0} \land \lnot F(X,Y)|_{y_i=1}$ is {\unsat} and a \textit{negative unate} if and only if $F(X,Y)|_{y_i=1} \land \lnot F(X,Y)|_{y_i=0}$ is {\unsat}~\cite{ACGKS18}.

Given a function vector $\langle \psi_1,\ldots,\psi_m \rangle$ for the vector of variables $\langle y_1, \ldots  y_m \rangle $ such that $\psi_i$ is the function corresponding to $y_i$, we say that there exists  a partial order  $\prec_{d}$ over the variables $\{y_1, \ldots y_m\}$ such that $y_i \prec_{d} y_j$ if $\psi_i$ depends on $y_j$. %We lift $\prec_{d}$ to sets of variables as follows: for $V_1, V_2 \subseteq Y$, $V_1 \prec_{d} V_2$ if $\forall y_i \in V_1, y_j \in V_2$ we have $y_i \prec_{d} y_j$.

%%Let us assume, we have given a function vector $\langle \psi_1,\cdots,\psi_m \rangle$ corresponding to the variables $\langle y_1,\cdots,y_m \rangle$, such that $\psi_i$ is function corresponding to $y_i$. We use $\prec$ to represent the dependency between two variables $y_i$ and $y_j$, if the function $\psi_i$ corresponding to $y_i$ depends on $y_j$, then $y_i \prec y_j$. We also use $\prec$ to represent the dependency between two sets of variables. Let $V_1$ and $V_2$ be any two subsets of $Y$, if for all $y_j \in V_2$, there does not exists a $y_i \in V_1$ such that $y_j \prec y_i$, then  $V_1 \prec V_2$. Note that, we have overloaded the definition of $\prec$ for the variables, and the set of variables.

%We define \textit{\totalorder} as an order of $Y$ variables. Let $y_i$ be the variable at $i^{th}$ index in \textit{\totalorder}, then $y_i$ $\prec \{{\totalorder}[i+1],\cdots, {\totalorder}[|Y|]\}$. %For a valid Skolem function vector there always exists at least one such {\totalorder}.

		In decision tree learning, a fraction of incorrectly assigned labels refer to the \textit{impurity}. We use Gini Index~\cite{Q86} as a measure of \textit{impurity} for a class label. The \textit{impurity decrease} at a node is the difference of its impurity to the mean of impurities of its children. The \textit{minimum impurity decrease} is a hyper-parameter used to control the maximum allowable impurity at the leaf nodes, thereby providing a lever for how closely the classifier fits the training data.

		Given a propositional formula $F(X,Y)$ and a weight function $W(\cdot)$ assigning non-negative weights to every literal, we refer to the \textit{weight} of a satisfying assignment $\sigma$, denoted as $W(\sigma)$, as the product of weights of all the literals appearing in $\sigma$, i.e., $W(\sigma) = \prod_{l \in \sigma} W(l)$. 
		A \emph{sampler} $\mathcal{A}(\cdot, \cdot)$ is a probabilistic generator that guarantees  $\forall \sigma \in \satisfying{F}, $ $\prob\left[\mathcal{A}(F, \bias) = \sigma\right] \propto W(\sigma)$.
		
%		The weight of a set of assignments $S$ is given by $W(S) = \sum_{\sigma \in S} W(\sigma)$~\cite{GSRM19}.
		 We use a function {\bias} that takes a mapping from a sequence of variables to the desired weights of their positive literals, and assigns corresponding weights to each of the positive literals. We use a simpler notation, {\bias}\texttt{(a,b)} to denote that positive literals corresponding to all universal variables are assigned a weight \texttt{a} and positive literals corresponding to all existential variables are assigned a weight \texttt{b}. For example, {\bias}(0.5, 0.9) assigns a weight of 0.5 to the positive literals of the universally quantified variables and 0.9 to the positive literals of the existentially quantified variables.
		%We overload the function {\bias}, it can also take a sequence of variables and a constant as an input, and it assigns the constant as weight to the positive literals of each of the variables.

\paragraph{\textbf{Problem Statement:}} Given a Boolean specification $F(X,Y)$ between set of inputs $X = \{x_1,\cdots,x_n\}$ and vector of outputs $Y=\langle y_1,\cdots,y_m\rangle$, the problem of \textit{Skolem function synthesis} is to synthesize a function vector $\Psi = \langle\psi_1(X),\cdots, \psi_m(X)\rangle$ such that $y_i \leftrightarrow \psi_i(X)$ and $\exists Y F(X,Y) \equiv F(X,\Psi)$. We refer to $\Psi$ as the \textit{Skolem function vector} and $\psi_i$ as the \textit{Skolem function} for $y_i$. 

A variable $y_i$ is called self-substituted variable, if the Skolem function $\psi_i$ corresponding to $y_i$ is set to $F(X,Y)|_{y_i=1}$~\cite{FTV16}.

 Given a formula $\exists Y F(X,Y)$ and a Skolem function vector $\Psi$, we refer to $E(X,Y,Y')$ as an \textit{error formula}~\cite{JSCTA15}, where $Y'=\{y'_1,\cdots,y'_{|Y|}\}$, and $Y' \neq Y$.
 \begin{equation}
 \label{eqn:errorformula}
 E(X,Y,Y') = F(X,Y) \land \lnot F(X,Y') \land (Y'\leftrightarrow \Psi)
 \end{equation}
 We use the following theorems from prior work:

			\begin{theorem}[\cite{JSCTA15}]
				$\Psi$ is a Skolem function if and only if $E(X,Y,Y')$ is {\unsat}.
			\end{theorem} 
			\begin{theorem}[\cite{ACGKS18}]
				\label{thm:unates}
				 If $y_i$ is positive(\textit{resp} negative) unate in $F(X,Y)$, then $\psi_i = 1$ (\textit{resp} $\psi_i = 0$) is the Skolem function for $y_i$.
			\end{theorem}
%			\begin{theorem}[\cite{FTV16}]
%			A variable $y_i$ is considered self-substituted, if $\psi_i= F(X,Y)|_{y_i=1}$ is the Skolem function for $y_i$, i.e. $\exists Y F(X,Y) \equiv F(X,Y|_{(y_i=F(X,Y)|_{y_i=1})})$.
%			\end{theorem}
\section{Related Work}
\label{sec:relatedwork}
The origins of the problem of Boolean functional synthesis traces back to Boole's seminal work~\cite{boole1847}, which was subsequently rigorously pursued, albeit focused on decidability, by Lowenheim and Skolem~\cite{L1910}. The complexity theoretic studies have shown that there exist instances where Boolean functional synthesis takes super polynomial time and was also shown that there exist instances for which polynomial size Skolem function vector does not suffice unless Polynomial Hierarchy (PH) collapses~\cite{ACGKS18}. 
%is Co-NP$^{\text{NP}}$-hard%

%The fundamental nature of Boolean functional synthesis has led to discovery of several practical applications ranging from program repair~\cite{JMF14}, program synthesis~\cite{SGF13}, cryptography~\cite{MM20}, logic synthesis~\cite{BJ11}, and the like. As such, there has been significant interest in the development of practical algorithms.
 Motivated by the success of the CEGAR (Counter-Example Guided Abstraction Refinement) approach in model checking, CEGAR-based approaches have been pursued in the context of synthesis as well, where the key idea is to use a Conflict-Driven Clause Learning (CDCL) SAT solver to verify and refine the candidate Skolem functions~\cite{AACKRS19,ACGKS18,ACJS17,JSCTA15}. 

Another line of work has focused on the representation of specification, i.e., $F(X,Y)$, in representations that are amenable to efficient synthesis for a class of functions. The early approaches focused on ROBDD representation building on the functional composition approach proposed by Balabanov and Jiang~\cite{BJ11}. Building on Tabajara and Vardi's ROBDD-based approach~\cite{TV17}, Chakraborty et al. extended the approach to factored specifications~\cite{CFTV18}. It is worth mentioning that factored specifications had earlier been pursued in the context of CEGAR-based approaches. Motivated by the success of knowledge compilation in the field of probabilistic reasoning, Akshay et al. achieved a significant breakthrough over a series of papers~\cite{ACGKS18,ACJS17,JSCTA15} to propose a new negation normal form, SynNNF~\cite{AACKRS19}. The generalization and a functional specification presented in SynNNF is amenable to efficient functional synthesis~\cite{AACKRS19}. Another line of work focused on the usage of {\em incremental determinization} to incrementally construct the Skolem functions~\cite{HSB14,JBSKW07,NPLSB12,R19,RTRS18}.

Several approaches have been proposed for the particular case when the specification, $\exists Y F(X,Y)$ is valid, i.e., $\forall X \exists Y F(X,Y)$ is True. Inspired by the sequential relational decomposition, Chakraborty et al.~\cite{CFTV18} recently proposed an approach focused on viewing each CNF clause of the specification consisting of {\em input and output} clauses and employing a {\em cooperation}-based strategy. The progress in modern CDCL solvers has led to an exploration of usage of heuristics for problems in complexity classes beyond NP. This has led to work on the extraction of Skolem functions from the proofs constructed for the formulas expressed as $\forall X \exists Y F(X,Y)$~\cite{BJ11,BJ12}.

%These approaches, however, fail to compute Skolem functions for the cases when the specification, $\exists Y F(X,Y)$, is realizable but not valid. 

%Priyanka : CADET computes skolem functions for  $\exists X \exists Y F(X,Y)$

The performance of Manthan crucially depends on its ability to employ constrained sampling, which has witnessed a surge of interest with approaches ranging from those based on hashing-based techniques~\cite{CMV14}, knowledge compilation~\cite{GSRM19,SGRM18}, augmentation of SAT solvers with heuristics~\cite{cryptominisat}. 

The recent success of machine learning has led to several attempts to the usage of machine learning in several related synthesis domains such as program synthesis~\cite{ABJMRSSSTU13}, invariant generation, decision-tree for functions in Linear Integer Arithmetic theory using pre-specified examples~\cite{FG19}, strategy synthesis for QBF~\cite{J18}.  
Use of data-driven approaches for invariant synthesis has been investigated in the ICE learning framework~\cite{ENDGM18,GLMN14,GNMR16} aimed with data about the program behavior from test executions, it proposes invariants by learning from data, checks for inductiveness and, on failure, extend the data by the generated counterexamples. 
%The primary contribution of their work was in handling inductiveness within the learning framework.
% In contrast to this, {\tool} uses a counterexample loop to successively ``repair" the proposed model instead of re-entering the learning phase. 
 The usage of proof-artifacts such as unsat cores has been explored in verification since early 2000s~\cite{GLST05} and in program repair in Wolverine~\cite{VR17}, while MaxSAT has been used in program debugging in~\cite{BPR16,JM11}.

\section{{\tool}: An overview}
\label{sec:overview}
\begin{figure}
		\begin{center}
			\includegraphics[scale=0.29]{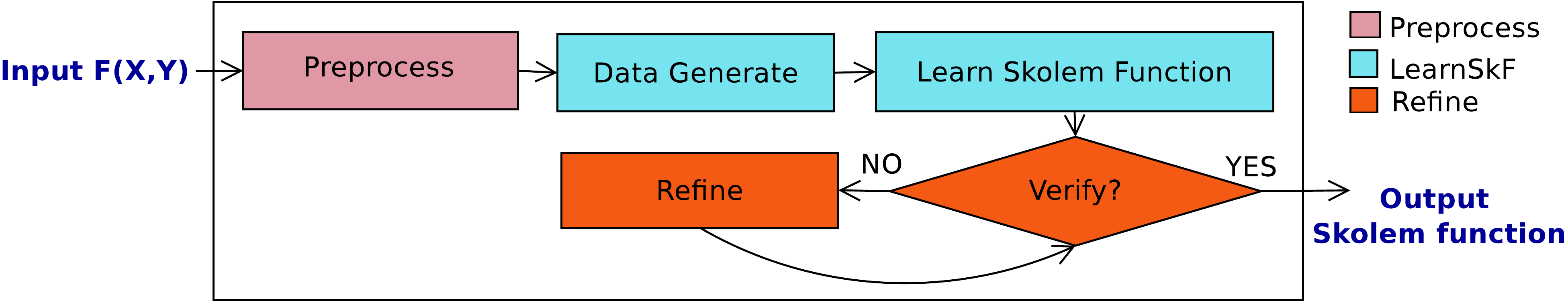}
			\caption{\label{fig:takneek_block_diagram} Overview of \tool}
		\end{center}
\end{figure}

In this section, we provide an overview of our proposed framework, {\tool}, before divulging into core algorithmic details in the following section. {\tool} takes in a function specification, represented as $F(X,Y)$, and returns a Skolem function vector $\Psi(X)$ such that $ \exists Y F(X,Y)$ $\equiv$ $F(X,\Psi(X))$. As shown in Figure~\ref{fig:takneek_block_diagram} {\tool} consists of following three phases:
\begin{enumerate}
	\item {\bf \phaseone} employs state-of-the-art pre-processing techniques on $F$ to compute a partial Skolem function vector. 
	\item {\bf \phasetwo} takes in the pre-processed formula and uses constrained samplers, and classification techniques to compute candidate Skolem functions for all the variables in $Y$. 
	\item {\bf \phasethree} performs verification and proof-guided refinement procedure wherein a {\sat} solver is employed to verify the correctness of candidate functions and a {\maxsat} solver in conjunction with a {\sat} solver is employed to refine the candidate functions until the entire candidate Skolem function vector passes the verification check.  
\end{enumerate}
% \tool{} consists of three phases. We call Phase one {\em \phaseone} as it preprocess the formula $F(X,Y)$. Phase called {\em \phasetwo}, consists of data generation and learning candidate skolem function using dependency-driven classifier. Phase three called, {\em \phasethree} performs counter example guided refinements(if necessary) to produce skolem functions.
We now provide a high-level description of different phases to highlight the technical challenges, which provides context for several algorithmic design choices presented in the next section. 
\subsection{Phase 1: \phaseone}
	{\phaseone} focuses on pre-processing of the formula to search for unates among the variables in $Y$; if $y_i$ is positive (resp. negative) unate, then $\psi_i = 1$(resp. $0$) suffices as a Skolem function. We employ the algorithmic routine proposed by Akshay et al.~\cite{ACGKS18} to drive this preprocessing.

\subsection{Phase 2: \phasetwo}

	{\phasetwo} views the problem of functional synthesis through the lens of machine learning where the learned machine learning model for classification of a variable $y_i$ can be viewed as a candidate Skolem function for $y_i$. We gather training data about the function's behavior by exploiting the progress in constrained sampling to sample solutions of $F(X,Y)$. Recall that $F(X,Y)$ defines a relation (and not necessarily a function) between $X$ and $Y$, and the machine learning techniques typically assume the existence of function between features and labels, necessitating the need for sophisticated sampling strategy as discussed below.  Moving on to features and labels, since we want to learn $Y$ in terms of $X$, we view $X$ as a set of features while assignments to $Y$ as a set of class labels. 
	
	The off-the-shelf classification techniques typically require that the size of training data is several times larger than the size of possible class labels, which would be prohibitively large for the typical problems involving more than thousand variables. To mitigate the requirement of large training data, we make note of two well-known observations in functional synthesis literature: (1) the Skolem function $\psi_i$ for a variable $y_i$ typically does not depend on all the variables in $X$, (2) A Skolem function vector $\Psi$ where $\psi_i$ depends on variable $y_j$ is a valid vector if the Skolem function $\psi_j$ is not dependent on $y_i$ (i.e., acyclic dependency), i.e., there exists a partial order $\prec_{d}$ over $\{y_1, \ldots y_m\}$.
	%Note that the latter condition is not an if-and-only-if as the Skolem functions are allowed to have conditional acyclic dependency conditioned on different assignments to $X$. In this work, we search for Skolem functions with stronger requirement of acylic dependency for all the assignments to $X$ and leave the relaxed variant to future work. 
	\par
	The above observations lead us to design an algorithmic procedure where we learn candidate Skolem functions as decision trees in an iterative manner, i.e., one $y_i$ at a time, thereby allowing us to constrain ourselves to the binary classification. The learned classifier can then be represented as the disjunction of all the paths from the root to the leaves in the learnt decision tree. We update the set of possible features for a given $y_i$ depending on the candidate functions generated so far, i.e., valuation of $X$ variables and $Y$ variables, which are not dependent on $y_i$. Finally, we compute the candidate Skolem function for $y_i$ as the disjunction of labels along edges for all the paths from the root to leaf nodes with label 1.
	Once, we have the candidate Skolem function vector $\Psi$, we obtain a valid linear extension, {\totalorder}, of the partial order $\prec_{d}$ in accordance to $\Psi$.
	 %Towards the end of {\phasetwo}, as discussed in section~\ref{preliminaries}, the {\totalorder} is constructed  in accordance to the candidate Skolem function vector. For a valid Skolem function vector, there always exists at least one such {\totalorder}.
	\par
	Before moving on to the next phase, we return to the formulation of sampling. The past few years have witnessed the design of uniform~\cite{CMV14,SGRM18}, and weighted samplers~\cite{GSRM19}, and one wonders what kind of sampler should we choose to generate samples for training data. A straightforward choice would be to perform uniform sampling over $X$ and $Y$, but the relational nature of specification, $F$, between $X$ and $Y$ offers interesting challenges and opportunities. Recall while $F$ specifies a relation between $X$ and $Y$, we are interested in a Skolem function, and we would like to tailor our sampling subroutines to allow discovery of Skolem functions with {\em small} description given the relationship between description and sample complexity. To this end, consider $X = \{x_1, x_2\}$ and $Y= \{ y_1\}$, and let $F:= (x_1 \vee x_2 \vee y_1)$. Note that $F$ has 7 solutions over $X \cup Y$, out of which $y_1=0$ appears in $3$ solutions while $y_1 = 1$ appears in 4. Also, note that there are several possible Skolem functions such as $y_1 = \neg (x_1 \wedge x_2)$. Now, if we uniformly sample solutions of $F$ over ${x_1, x_2, y_1}$, i.e. {\bias}(0.5, 0.5), we would see (almost) equal number of samples with $y_1 = 0$ and $y_1 = 1$. A closer look at $F$ reveals that it is possible to construct a Skolem function by knowing that the only case where $y_1$ cannot be assigned $0$ is when $x_1 = x_2 =0$. To encode this intuition, we propose a novel idea of collecting samples with weighted sampling, i.e., {\bias}(0.5, q) where $q$ is chosen in a multi-step process of first drawing a small set of samples with both $q = 0.9$ and $q=0.1$, and then drawing rest of the samples by fixing the value of $q$ following analysis of an initial set of samples. To the best of our knowledge, this is the first application of weighted sampling in the context of synthesis, and our experimental results point to several interesting avenues of future work.
	
%	In {\phasetwo} phase, {\tool} generates data and uses it to learn candidate skolem function. {\tool} considers the problem of generating candidate skolem function as binary classification problem over samples of $\exists Y F(X,Y)$ and uses valuation of universally quantified variables as features and valuation of existentially quantified variables as label. {\tool} uses state-of-the-art decision tree technique to do binary classification over samples to learn candidate skolem functions.
%	\par
%	{\tool} creates decision tree for every $Y$ variable to find candidate skolem function. {\tool} needs the different feature set for every $Y$ variable to avoid cyclic dependencies as it does not consider any predefined order of variables. To understand better, let us assume {\tool} is creating decision tree for a $y_j$ of $Y$.     
%	The feature set for $y_j$ includes valuation of $X$ variables and $Y$ variables, which are not dependent on $y_j$. If a variable $y_k$ is dependent on $y_j$, then in decision tree generation of $y_j$, valuation of $y_k$ is not considered as feature. Candidate skolem function for $y_j$ is disjunction of all the paths where leaf node has label $1$. 
%	\par
%	One can generate samples uniformly at random to do classification, but unlike the standard classification problem, where we have a unique label for each assignment of features, in our case, we can have a different label, that is assignment of $Y$ variable for the same assignments of features($X$ variables). We use weighted samples to overcome the issue. 
%		\input{chapters/specialsamples.tex}  
\subsection{Phase 3: \phasethree}
	The candidate Skolem functions generated in {\phasetwo} may not always be the actual Skolem functions. Hence, we require a \textit{verification} check to see if candidate Skolem functions are indeed correct; if not, the generated counterexample can be used to {\em repair} it. The verification query constructs an \textit{error formula} $E(X,Y,Y')$(Formula \ref{eqn:errorformula}): if unsatisfiable, the candidate Skolem function vector is indeed a Skolem function vector and the procedure can terminate; else, when $E(X,Y,Y')$ is {\sat}, the solution of $E(X,Y,Y')$ is used to identify and refine the erring functions among the candidate Skolem function vector.
	
	In contrast to prior techniques that apply Shannon expansion or self-substitution, the refinement strategy in {\tool} is guided by the view that the candidate function vector from the {\phasetwo} phase is  \textit{almost correct}, and hence, attempts to identify and apply a series of {\em minor} repairs to the erring functions to arrive at the correct Skolem function vector. To this end, {\tool} uses two key techniques: {\it fault localization} and {\it repair synthesis}. 
	Let us assume that $\sigma$ is a satisfying assignment of $E(X,Y,Y')$ and referred to as counterexample for the current candidate Skolem function vector $\Psi$. 
	
		\paragraph{\bfseries Fault Localization} 
        %In order to identify the {\it minimal} to repair for the counterexample $\sigma$, {\tool} attempts to identify the smallest number of Skolem functions (correspondingly $Y$ variables) whose outputs must undergo a change for the formula to behave correctly on $\sigma$; in other words, it attempts to find the \textit{maximum} number of Skolem function (correspondingly $Y$ variables) that can be allowed to retain their current output on $\sigma$ while satisfying the formula. {\tool} encodes this problem as a partial MaxSAT query with $F(X,Y) \land (X \leftrightarrow \sigma[X])$ as a hard constraint and $(Y \leftrightarrow \sigma[Y'])$ as soft constraints. All $Y$ variables whose valuation constraint ($(Y \leftrightarrow \sigma[Y'])$) does not hold in the MAXSAT solution are identified as erring Skolem functions that need to be repaired.
        In order to identify the initial candidates to repair for the counterexample $\sigma$, {\tool} attempts to identify a small number of Skolem functions (correspondingly $Y$ variables) whose outputs must undergo a change for the formula to behave correctly on $\sigma$; in other words, it makes a best-effort attempt to ensure that most of the Skolem functions (correspondingly $Y$ variables) can retain their current output on $\sigma$ while satisfying the formula. {\tool} encodes this problem as a partial {\maxsat} query with $F(X,Y) \land (X \leftrightarrow \sigma[X])$ as a hard constraint and $(Y \leftrightarrow \sigma[Y'])$ as soft constraints. All $Y$ variables whose valuation constraint $(Y \leftrightarrow \sigma[Y'])$ does not hold in the {\maxsat} solution are identified as erring Skolem functions that may need to be repaired.
		\paragraph{\bfseries  Repair Synthesis}
%		\todo{Subhajit Sir, please review this}
        Let $y_k$ be the variable corresponding to the erring  function, $\psi_k$, identified in the previous step. To synthesize a repair for the function, {\tool} applies a proof-guided strategy: it constructs a formula $G_k(X,Y)$, such that if $G_k(X,Y)$ is unsatisfiable then $\psi_k$ must undergo a change. The {\unsatcore} of $G_k(X,Y)$ provides a \textit{reason} that explains the discrepancy between the specification and the current Skolem function.
		%(Formula \ref{eqn:refinement sat query}). 
			\begin{multline}
			\label{eqn:refinement sat query}
			G_{k}(X,Y) = (y_k \leftrightarrow \sigma[y'_{k}]) \land F(X,Y) \land (X \leftrightarrow \sigma[X]) \land (\hat{Y} \leftrightarrow \sigma[\hat{Y}])
			\\
			\mathrm{where}\: \hat{Y} \subset Y \; \mathrm{and} \; \hat{Y} =  \{ \totalorder[index(y_k)+1],\cdots,\totalorder[|Y|]\} \ \;
			\end{multline}
%		\begin{multline}
%  		 	\label{eqn:refinement sat query}
%			G_{k}(X,Y) = (y_k \leftrightarrow \sigma[y'_{k}]) \land F(X,Y) \land (X \leftrightarrow \sigma[X]) \land (\hat{Y} \leftrightarrow \sigma[\hat{Y}])
%			\\
%			\mathrm{where}\: \hat{Y} \subset Y \;s.t.\; \forall y_j \in \hat{Y}: \psi_{k} \:\mathrm{depends\:on}\: y_{j}
%		\end{multline}

 {\tool} uses the {\unsatcore} to constructs a \textit{repair formula}, say~$\beta$, as a conjunction over literals in the unsatisfiable core; if $\psi_k$ is {\it true} with the current valuation of $X$ and $\hat{Y}$, {\tool} updates the function $\psi_k$ by conjoining it with the negation of repair formula ($\psi_k \gets \psi_k \land \lnot \beta$); otherwise, {\tool} updates the function $\psi_k$, by disjoining it with the repair formula ($\psi_k \gets \psi_k \lor  \beta)$.
%		\begin{itemize}
%            \item If $G(X,Y)$ is unsatisfiable, it implies that there exists a reason within $\hat{Y}$ that can explain the discrepency between the specification and the current skolem functions. In this case, {\tool} constructs a \textit{repair formula}, say~$\beta$, as a conjunction over literals in the unsatisfiable core from the proof of infeasibility. If the erring value of the function is {\it true}, we strengthen the function $\psi_k$ by conjoining it with the repair formula ($\psi_k \gets \psi_k \land \lnot \beta$); otherwise, we weaken the  function $\psi_k$  by disjoining it with the negation of the repair formula ($\psi_k \gets \psi_k \lor  \beta)$.
%            \item If $G(X,Y)$ is satisfiable, it implies that the current Skolem function $\psi_k$ of $y_k$ is consistent with valuation of $X$ and $\hat{Y}$, and the reason of the discrepensy lies outside $\hat{Y}$ (or the specification is already met due to corrections to other Skolem functions in the previous loop iterations). In such a case, {\tool} attempts to find $y_t \notin \hat{Y}$, whose Skolem function $\psi_t$ is not consistent with current $\psi_k$, and it adds $\psi_t$ to the list of erring Skolem functions.
%		\end{itemize}
	%The ordering of the variable is important; $G(X,Y)$ does not constrain $Y$ variables that depend on $y_k$ to avoids cyclic dependencies between variables.

    \subsubsection{Self-substitution for poorly learnt functions} Some Skolem functions are difficult to learn through data.  In our implementation, the corresponding variables escape the {\phasetwo} phase with poor candidate functions, thereby requiring a long sequence of incremental repairs for convergence. To handle such scenarios, we make the following observation: though synthesizing Skolem functions via self-substitution\cite{FTV16} can lead to an exponential blowup in the worst case, it is inexpensive if the number of variables synthesized via this technique is small. We use this observation to quickly synthesize a Skolem function for an erring variable if we detect its candidate function is poor (detected by comparing the number of times it enters refinement against an empirically determined threshold). Of course, this heuristic does not scale well if the number of such variables is large; in our experiments, we found less than 20\% of the instances solved required self-substitution, and for over 75\% of these instances, only one variable needed self-substitution. We elaborate more on the empirical evidence on the success of this heuristic in section~\ref{sec:experiments}. A theoretical understanding of the learnability of Boolean functions from data seems to be an interesting direction for future work.
%	and we conjecture that this may lead to interesting connections with respect to complexity theoretic hardness of learning of skolem functions. 
	%The primary reason behind repeated refinements was the lack of good guesses of candidate skolem functions. The decision trees for those variables were not able to learn candidate skolem function as it had just one node.
	%To overcome the issue, we used self substitution\cite{cegar}. We discuss the impact of self substitution in section \ref{experimentsection}.
	%{\tool} identifies an existentially quantified variables for which the decision tree did not learn, call them the ``suspicious variable". \tool{} self substitutes a $y_i$ of $Y$ only if a $y_i$ is a suspicious variable and it occurs in refinement process for more than $10$ times. 
	%\newline
	%\noindent
	%At the end to give skolem function vector as a function of only $X$ variables, {\tool} sequentially substitutes $Y$ variables in skolem function $\psi_j$ of $y_j$ with corresponding skolem function.

\section{{\tool}: Algorithmic Description}\label{sec:algorithm}
\begin{algorithm}
\footnotesize
\DontPrintSemicolon
\LinesNumbered
\SetAlgoLined
%\KwIn{$\exists Y F(X,Y)$ where $X=\{x_1,x_2,\cdots,x_n\}, Y=\{y_1,y_2,\cdots,y_m\}$}
%\KwOut{Skolem function $\Psi = (\psi_1,\cdots,\psi_m)$}

$\Psi,U \gets$ {\preprocess}$(F(X,Y))$   \label{mainalgo:line:preprocess}\; 

$\Sigma \gets$ {\getsamples}($F(X,Y)$) \label{getsample} \;

$D \gets \emptyset$\;

\ForEach { $y_j \in Y \setminus U$ \label{mainalgo:line:phase1start}}
{
	$\psi_j,D \gets$ {\candidateskf}($\Sigma,F(X,Y),y_j,D$)\; \label{mainalgo:line:phase1end}
} 

$\totalorder \gets$ {\findorder}($D$) \label{findorder} \;

\Repeat{ret = UNSAT}
{
	$E(X, Y, Y') \gets F(X,Y) \land \lnot F(X,Y') \land (Y'\leftrightarrow \Psi) $
	\label{mainalgo:line:refinestart}\;
	
	$ret, \sigma \gets$ {\checksat}($E(X, Y, Y')$) \label{checksat}\; 

	\If{ret = SAT}{
		$\Psi \gets {\refine}(F(X,Y),\Psi,\sigma,{\totalorder})$  \label{mainalgo:line:refineend}\;}
 }

$\Psi \gets$ {\substitute}($F(X,Y),\Psi,{\totalorder}$)\label{mainalgo:line:reversesubstitute}\;

\Return $\Psi$ \;
\caption{{\tool}($F(X,Y)$)}
\label{main_algo}
\end{algorithm}

In this section, we present a detailed algorithmic description of {\tool}, whose pseudocode is presented in Algorithm~\ref{main_algo}. {\tool} takes in a formula $F(X,Y)$ as input and returns a Skolem vector $\Psi$. 
The algorithm starts off by preprocessing (line~\ref{mainalgo:line:preprocess}) the formula $F(X,Y)$ to get the unates ($U$) and their corresponding Skolem functions ($\Psi$). Next, it invokes the sampler (line~\ref{getsample}) to collect a set of samples$(\Sigma)$ as training data for the learning phase.

For each of the existential variables that are not unates, {\tool} attempts to learn candidate Skolem functions (lines~\ref{mainalgo:line:phase1start}-\ref{mainalgo:line:phase1end}). %To avoid the cyclic dependencies among the variables in the generated candidate Skolem functions%
 To generate a variable order, {\candidateskf} uses a collection of sets $d_1,\cdots,d_{|Y|} \in D$, such that $y_i \in d_j$ indicates that $y_j$ depends on $y_i$. Next, the {\findorder} routine (line~\ref{findorder}) construct {\totalorder} of the $Y$ variables in accordance to the dependencies in $D$.
%To get order of dependencies between $Y$ variables ($Yorder$, {\toolalgo} calls {\findorder} with set $D$. 
The verification and refinement phase (line~\ref{mainalgo:line:refinestart}) commences by constructing the error formula and launching the verification check (line~\ref{checksat}). If the error formula is satisfiable, the counterexample model ($\sigma$) is used to refine the formula. Once the verification check is successful, the refinement phase ends and the subroutine {\substitute} is invoked to recursively substitute all $y_i \in Y$ appearing in Skolem functions with their corresponding Skolem functions such that only $X$ variables entirely describe all Skolem functions. The strict variable ordering enforced above ensures that {\substitute} always succeeds and does not get stuck in a cycle. Finally, the Skolem function vector $\Psi$ is returned.

It is worth noting that {\tool} can successfully solve an instance without having to necessarily execute all the phases. In particular, if $U = Y$, then {\tool} terminates after {\phaseone} (i.e., line~\ref{mainalgo:line:preprocess}). Similarly, if the {\checksat} return {\unsat} during the first iteration of loop (lines~\ref{mainalgo:line:refinestart}--\ref{mainalgo:line:refineend}), then {\tool} does not invoke {\refine}.
We now discuss each subroutine in detail.
%\begin{description}
%	\item [Preprocess]
	\paragraph{\preprocess:}
		\begin{algorithm}
	\footnotesize
	\DontPrintSemicolon
	\SetAlgoLined	
	$U \gets \emptyset$\;
	
	\ForEach {$y_j \in Y$}
	{ 
		$ret_{pos}, \rho_{pos} \leftarrow {\checksat}(F(X,Y)|_{y_j=0} \land \lnot F(X,Y)|_{y_j=1})$ \label{algo-preprocess-line-positive-start}\;
		\If {$ret_{pos}$ = \unsat } 
		{
			
			$U \gets U \cup y_j$\;
			$F(X,Y) \gets F(X,Y)|_{y_j=1}$ \;
			$\psi_j \gets 1$ ~~~\tcc*[h]{$y_j$ is positive unate} \; \label{algo-preprocess-line-positive-end}
		}
		\Else{ 
			$ret_{neg}, \rho_{neg} \leftarrow {\checksat}(F(X,Y)|_{y_j=1} \land \lnot F(X,Y)|_{y_j=0})$ \label{algo-preprocess-line-negative-start}\;
			\If{$ret_{neg}$ = \unsat  } 
			{
				
				$U \gets U \cup y_j$\;
				$F(X,Y) \gets F(X,Y)|_{y_j=0}$ \;
				$\psi_j \gets 0$ ~~~\tcc*[h]{$y_j$ is negative unate} \; \label{algo-preprocess-line-negative-end}
			}
		}
	}
	\Return $\Psi,U$
	\caption{{\preprocess}($F(X,Y)$)}
	\label{preprocessing_F}
\end{algorithm}
			We perform the pre-processing step as described in~\cite{ACGKS18}, which performs SAT queries on the formulas constructed as specified in Theorem~\ref{thm:unates}. 
		 As described in Algorithm \ref{preprocessing_F}, 	{\preprocess} sequentially detects unate for every $y_j$ of $Y$ variable. {\preprocess} first checks for the satisfiability of positive unate formula (line \ref{algo-preprocess-line-positive-start}) for $y_j$. If $y_j$ is a positive unate, then $\psi_j$ is $1$ (line \ref{algo-preprocess-line-positive-end}), and $y_j$ is added to set $U$. If $y_j$ is not a positive unate, then {\preprocess} checks for the satisfiability of negative unate formula (lines \ref{algo-preprocess-line-negative-start}). If $y_j$ is a negative unate, then $\psi_j$ is $0$ (line \ref{algo-preprocess-line-negative-end}), and $y_j$ is added to set $U$.

	\paragraph{\getsamples:} 
			\begin{algorithm}
	\footnotesize
	\DontPrintSemicolon
	\SetAlgoLined
	$\Sigma_1 \gets $ \adasampler($F(X,Y),500,0.5,0.9$)\label{getsamples_algo_line_bias0.9}\;
	$\Sigma_2 \gets $ \adasampler($F(X,Y),500,0.5,0.1$)\label{getsamples_algo_line_bias0.1}\;
	\ForEach {$y_j \in Y$}
	{
		$m_j \gets $ Count($\Sigma_1 \cap (y_j =1))/500$\label{getsamples_algo_line_calculate_mj}\;
		$n_j \gets $ Count($\Sigma_1 \cap (y_j =0))/500$ \label{getsamples_algo_line_calculate_nj}\;
		\If  {$(0.35 < m_j < 0.65) \land (0.35 < n_j < 0.65)$ }
		{
			$q_j \gets m_j$\;
		}
		\Else{
			$q_j \gets 0.9$\;
		}
	}
	$\Sigma \gets  $ \adasampler($F(X,Y),0.5,q$)\label{getsamples_algo_line_biasq}\;
	\Return $\Sigma$\;
	\caption{{\getsamples}($F(X,Y)$)}
	\label{getsamples_algo}
\end{algorithm}
		{\getsamples}, presented in Algorithm~\ref{getsamples_algo},  takes $F(X,Y)$ as input and returns, $\Sigma$, a subset of satisfying assignments of $F(X,Y)$. {\getsamples} first generates $500$ samples each with {\bias}$(0.5,0.9)$(line \ref{getsamples_algo_line_bias0.9}), and with {\bias}$(0.5,0.1)$(line \ref{getsamples_algo_line_bias0.1}). Then, {\getsamples} in line~\ref{getsamples_algo_line_calculate_mj}, calculates $m_j$ for all $y_j$, $m_j$ is a ratio of number of samples with $y_j$ being $1$ to the total number of samples, i.e. 500. Similarity, in line~\ref{getsamples_algo_line_calculate_nj}, it calculates $n_j$ for all $y_j$, $n_j$ is a ratio of number of samples with $y_j$ being $0$ to the total number of samples. Finally, {\getsamples} generates required number of samples with {\bias}$(0.5,q)$; for a $y_j$, $q$ is $m_j$ if both $m_j$ and $n_j$ are in range $0.35$ to $0.65$, else $q$ is $0.9$.

	\paragraph{\candidateskf:}
			\begin{algorithm}[t]
\footnotesize
\DontPrintSemicolon
\LinesNumbered
\SetAlgoLined
%\KwIn{$S, F(X,Y), y_j,D$}
%\KwOut{$\psi_j, D$}

$featset \gets X$    \;

\ForEach {$y_k \in Y\setminus y_j$ \label{algo-decisiontree-line-feature-start} }
{ 
    \If {$y_j \notin d_k$ }
   		{
   			$featset \gets featset \cup y_k$ \tcc*[h]{if $y_k$ is not dependent on $y_j$}
   		}
}\label{algo-decisiontree-line-feature-end}

$feat,lbl \gets$ $\Sigma_{\downarrow featset}, \Sigma_{\downarrow y_j}$\;

$t \gets$ {\createdt}$(feat,lbl)$\;
 
\ForEach { n $\in$  $\mathrm{LeafNodes}$(t)\label{algo-decisiontree-line-skf-start}}
{
	\If  {{\Label}(n) = 1 \label{algo-decisiontree-tree}}
		{
			$\pi \gets$ {\Path}$(t,root,n)$\;
			$\psi_j \gets \psi_j \lor \pi$\;
		}
}\label{algo-decisiontree-line-skf-end}

\ForEach {$y_k \in \psi_j$}
{
	$d_j \gets d_j \cup y_k \cup d_k$ \label{algo-decisiontree-line-dependent}
}

\Return $\psi_j,D$\;

\caption{{\candidateskf}$(\Sigma,F(X,Y),y_j,D)$}
\label{decision_tree_skf_algo}
\end{algorithm}

			{\candidateskf}, presented in Algorithm~\ref{decision_tree_skf_algo}, assumes access to following three subroutines:
			\begin{enumerate}[topsep=0pt]
%					\item {\isleaf} takes a node of the tree $t$ as input and returns true if node is a leaf node otherwise it returns false.		
                \item{\createdt}  takes the feature and label sets as input (training data) and returns a decision tree $t$. We use the ID3 algorithm~\cite{Q86} to construct a decision tree $t$ where the internal node of $t$ represents a feature on which a decision is made, the branches represent partitioning of the training data on the decision, and the leaf nodes represent the classification outcomes (i.e class labels). The ID3 algorithm iterates over the training data, and in each iteration, it selects a new attribute to extend the tree by a new decision node: the selected attribute is one that causes the maximum drop in the impurity of the resulting classes; we use Gini Index~\cite{Q86} as the measure of impurity. The algorithm, then, extends the tree by the selected decision and continues extending building the tree. The algorithm terminates on a path if either it exhausts all attributes for decisions, or the impurity of the resulting classes drop below a (user-specified) impurity decrease parameter.

                \item {\Label} takes a leaf node of the decision tree as input and returns the class label corresponding to the node.
					\item {\Path} takes a tree $t$ and two nodes of $t$ (node $a$ and node $b$) as input and outputs a conjunction of literals in the path from node $a$ to node $b$ in $t$.
			\end{enumerate}
            As we seek to learn Boolean functions, we employ binary classifiers with class labels \texttt{0} and \texttt{1}. {\candidateskf} shows our algorithm for extracting a Boolean function from the decion trees: lines \ref{algo-decisiontree-line-feature-start}-\ref{algo-decisiontree-line-feature-end} find a feature set ({\em featset}) to predict $y_j$. The feature set includes all $X$ variables and the subset of $Y$ variables that are not dependent on $y_j$. {\candidateskf} creates decision tree $t$ using samples $\Sigma$ over the feature set. Lines~\ref{algo-decisiontree-line-skf-start}-\ref{algo-decisiontree-line-skf-end} generate candidate Skolem function $\psi_j$ by iterating over all the leaf nodes  of $t$. In particular, if a leaf node is labeled with~\texttt{1}, the candidate function is updated by disjoining with the formula returned by subroutine {\Path}.
			{\candidateskf} also updates $d_j$ in $D$, $d_j$ is set of all $Y$ variables on which, $y_j$ depends. If $y_j$ depends on $y_k$, then by transitivity $y_j$ also depends on $d_k$; in line \ref{algo-decisiontree-line-dependent}, {\candidateskf} updates~$d_j$ accordingly.

	\paragraph{\findorder:}
		{\findorder} takes $D$ as an input to output a valid linear extension of the partial order $\prec_{d}$ defined over $\{y_1, \ldots y_m\}$ with respect to the candidate Skolem function vector $\Psi$. 

	\paragraph{\refine:} 
			% Refinement algorithm
\begin{algorithm}
	\footnotesize
\DontPrintSemicolon
%\LinesNumbered
\SetAlgoLined

%\KwIn{$F(X,Y)$,$\Psi,\sigma,D$}
%\KwOut{$\Psi$}

$H \gets F(X,Y) \land (X \leftrightarrow \sigma[X])$; $S \gets (Y \leftrightarrow \sigma[Y'])$\;

$Ind \gets$ {\maxsatlist}($H,S$)\; \label{algo:refine:line:maxsat}
\ForEach {$y_k \in Ind$}
{
%% check_sat call
	 \label{algo:refine:line:repair-start}
	$\hat{Y} \gets  \{ \totalorder[index(y_{k})+1],\cdots,\totalorder[|Y|] \}  $ \label{algo:refine:line:dependentset}\;

	\If{{\checksub}$(y_k)$}
	{
		$\psi_k \gets$ {\selfsub}$(F(X,Y),y_k,Y \setminus \hat{Y})$\;
	}
	\Else
	{

  		$G_k \gets  (y_k \leftrightarrow \sigma[y'_k]) \land F(X,Y) \land  (X \leftrightarrow \sigma[X]) \land (\hat{Y} \leftrightarrow \sigma[\hat{Y}]) $ \;

  		$ret, \rho \gets$ {\checksat}$(G_k)$ \label{algo:refine:line:checksat}\;

		%% if check_sat call is unsat
		\If { $ret = UNSAT$}
		{
 			$C \gets$ {\findcore}$(G_k)$ \label{algo:refine:line:unsat}\;
		
			$\beta \gets \underset{ l \in  C}{\bigwedge} ite((\sigma[l]=1),l,\lnot l)$ \label{refinement formula}\;

			$\psi_k \gets ite((\sigma[y'_k]=1),\psi_k \land \lnot \beta, \psi_k \lor \beta) $ \label{refined skolemfunction}\;

		}
		%% if check_sat call is sat
		\Else
		{

			\ForEach{ $y_t \in Y \setminus \hat{Y} $\label{algo:refine:line:sat-start}}
			{
				\If{ $\rho[y_t] \neq \sigma[y'_t] $}
				{
					$Ind \gets  Ind.Append(y_t)$\;
				}
			}
			$\sigma[y_k] \leftarrow \sigma[y'_k]$\;
		}\label{algo:refine:line:sat-end} 
	}
}
\Return $\Psi$\; \label{algo:refine:line:repair-end} 
\caption{{\refine}($F(X,Y),\Psi,\sigma,{\totalorder}$)}
\label{refine_algo}
\end{algorithm}

			{\refine} is invoked with a counterexample $\sigma$. {\refine} first performs \textit{fault localization} to find the initial set of erring candidate functions; to this end, it calls the {\maxsatlist} subroutine  (line  \ref{algo:refine:line:maxsat}) with $F(X,Y) \land (X \leftrightarrow \sigma[X])$ as hard-constraints and $(Y \leftrightarrow \sigma[Y])$ as soft-constraints. {\maxsatlist} employs a {\maxsat} solver to find the solution that satisfies all the hard constraints and maximizes the number of satisfied soft constraints, and then returns a list ({\em Ind}) of $Y$ variables such that for each of the variables appearing in  ({\em Ind})  the corresponding soft-constraint was not satisfied by the optimal solution returned by {\maxsat} solver.
			
			Since candidate Skolem function corresponding to the variables in {\em Ind} needs to refine, {\refine} now attempts to synthesize a repair for each of these candidate Skolem functions.	Repair synthesis loop (lines~\ref{algo:refine:line:repair-start}--\ref{algo:refine:line:repair-end}) starts off by collecting the set of $Y$ variables, $\hat{Y}$, on which $y_k$ of {\em Ind} can depend on as per the ordering constraints (line \ref{algo:refine:line:dependentset}). Next, it invokes the subroutine {\checksub}, which returns True if the candidate function corresponding to $y_k$ has  been refined more than a chosen threshold times (fixed to 10 in our implementation), and the corresponding decision tree constructed during execution {\candidateskf} has exactly one node.
%			 it calls {\checksub} to check whether $y_k$ is such a candidate. The Skolem function of an existentially quantified variable is constructed via self-substitution if the variable is a {\em suspicious} variable and is in refinement for more than $10$ times; an existentially quantified variable is called {\em suspicious} variable if the number of nodes in the decision tree generated to find candidate skolem function is $1$ (showing that the learning phase failed to learn a good model corresponding to this variable).		
			If {\checksub} returns true, {\refine} calls {\selfsub} to perform self-substitution. {\selfsub} takes a formula $F(X,Y)$, an existentially quantified variable $y_k$ and a list of variables which depends on $y_k$ and performs self substitution of $y_k$ with constant $1$ in the formula $F(X,Y)$\cite{JSCTA15}.
			
			If {\checksub} returns false, {\refine} attempts a proof-guided repair for $y_k$. {\refine} calls {\checksat} in line \ref{algo:refine:line:checksat} on $G_{k}$, which corresponds to formula \ref{eqn:refinement sat query}: if $G_{k}$ is {\sat}, then {\checksat} returns a satisfying assignment$(\rho)$ of $G_{k}$ in $\sigma$, else {\checksat} returns unsatisfiable in the result, $ret$. 
			%\par 
			%If $ret$ is {\unsat}, then the Skolem function value of $\psi_k$  with the current valuations of $X$ and $\hat{Y}$ is not same as $\sigma[y_i]$, and {\tool} attempts to refine $\psi_k$; else  value of $\psi_k$ is same as $\sigma_[y_k]$ with the current  valuations of $X$ and $\hat{Y}$, {\tool} does not refine $\psi_k$ but attempts to update its {\em Ind} list with $y_t \notin \hat{Y}$:
%            \item If $G(X,Y)$ is unsatisfiable, it implies that there exist a reason within $\hat{Y}$ that can explain the discrepency between the specification and the current skolem functions. In this case, {\tool} constructs a \textit{repair formula}, say $\beta$, as a conjunction over literals in the unsatisfiable core from the proof of infeasibility. If the erring value of the function is {\it true}, we strengthen the function $\psi_k$ by conjoining it with the repair formula ($\psi_k \gets \psi_k \land \lnot \beta$); otherwise, we weaken the  function $\psi_k$  by disjoining it with the negation of the repair formula ($\psi_k \gets \psi_k \lor  \beta)$.
%            \item If $G(X,Y)$ is satisfiable, it implies that the current Skolem function $\psi_k$ of $y_k$ is consistent with valuation of $X$ and $\hat{Y}$, and the reason of the discrepensy lies outside $\hat{Y}$ (or the specification is already met due to corrections to other Skolem functions in the previous loop iterations). In such a case, {\tool} attempts to find $y_t \notin \hat{Y}$, whose Skolem function $\psi_t$ is not consistent with current $\psi_k$, and it adds $\psi_t$ to the list of erring Skolem functions.
            \begin{enumerate}
			\item If $ret$ is {\unsat}, we proceed to refine $\psi_k$ such that for $\psi_k(X\mapsto \sigma[X],\hat{Y}\mapsto \sigma[\hat{Y}]) = \sigma[y_k]$. Ideally, we would like to apply a refinement that generalizes to potentially other counter-examples, i.e. solutions of $E(X,Y,Y')$. To this end, {\refine} calls {\findcore}  with $G_{k}$; {\findcore} returns the list of variables ($C$) that occur in the clauses of {\unsatcore} of $G_{k}$. Accordingly, the algorithm constructs a \textit{repair formula} $\beta$ as a conjunction of literals in $\sigma$ corresponding to variables in $C$   (line~\ref{refinement formula}).
            	If $\sigma[y'_k]$ is $1$, then $\psi_k$ is $\psi_k$ with conjunction of negation of $\beta$ and if $\sigma[y'_k]$ is $0$, then $\psi_k$ is $\psi_k$ with disjunction of $\beta$.  
            	
%                \item If $ret$ is {\unsat}, it implies that there exist a reason within $\hat{Y}$ that can explain the discrepancy between the specification and the current Skolem functions. In such a case (line \ref{algo:refine:line:unsat}) {\refine} calls {\findcore} with $G_{k}$; {\findcore} returns the list of variables ($C$) that occur in the clauses of the unsatisfiability proof of $G_{k}$. Accordingly, the algorithm constructs a \textit{repair formula} $\beta$ as a conjunction of literals in $C$(Line~\ref{refinement formula}).
%                If $\sigma[y'_k]$ is $1$, then $\psi_k$ is $\psi_k$ with conjunction of negation of $\beta$ (Line~\ref{refined skolemfunction}) and if $\sigma[y'_k]$ is $0$, then $\psi_k$ is $\psi_k$ with disjunction of $\beta$(Line~\ref{refined skolemfunction}). 
					
                \item If $ret$ is {\sat} and $\rho$ is a satisfying assignment of $G_{k}$, then there exists a Skolem function vector such that the value of $\psi_k$ agrees with $\sigma[y_k]$ for the valuation of $X$ and $\hat{Y}$ set to $\sigma[X]$ and $\sigma[\hat{Y}]$. However, for any $y_t \in Y \setminus \hat{Y}$ if $\sigma[y'_t] \neq \rho[y'_t]$, then for such a $y_t$, the Skolem function corresponding to $y_t$ may need to refine . Therefore, {\refine} adds $y_t$ to list of candidates to refine, $Ind$. Note that since $\sigma \models E(X,Y,Y')$, there exists at least one iteration of the loop (lines~\ref{algo:refine:line:repair-start}--~\ref{algo:refine:line:sat-end}) where $ret$ is {\unsat}.
                
%                 that the value of current Skolem function $\psi_k$ of $y_k$ is same as $\sigma[y_k]$  for valuations of $X$ and $\hat{Y}$ under $\sigma$ bwe may need to refine $y_t$ in $Y \setminus \hat{Y}$ if $y'_t$ in satisfying assignment of $E(X,Y,Y')$ (Formula \ref{eqn:errorformula}) is different than valuation of $y_t$ in satisfying assignment of $G_{k}$.
%                
%                , and the reason of the discrepancy lies outside $\hat{Y}$ (or the specification is already met due to corrections to other Skolem functions in the previous loop iterations). In this case, {\refine} (lines \ref{algo:refine:line:sat-start}-\ref{algo:refine:line:sat-end}) attempts to find $y_t$ in $Y \setminus \hat{Y}$ such that the valuation  of $y'_t$ in satisfying assignment of $E(X,Y,Y')$ (Formula \ref{eqn:errorformula}) is different than valuation of $y_t$ in satisfying assignment of $G_{k}$. {\refine} adds $y_t$ to list of candidates to refine, $Ind$.
			\end{enumerate}
		\paragraph{\substitute:}
			\begin{algorithm}
	\footnotesize
	\DontPrintSemicolon
	\SetAlgoLined
	%\KwIn{$\Psi,D$}
	%\KwOut{$\Psi$}
	\ForEach {$y_j \in Y$}
	{
		$\hat{Y} \gets  \{\totalorder[index(y_{j})+1],\cdots,\totalorder[|Y|] \}  $\;
		\ForEach {$y_i \in \hat{Y}$}
		{
			$\psi_j \gets \psi_j|_{y_i \leftrightarrow \psi_i}$\;
			%d_j \gets d_j \cup d_k$
		}
	}
	\Return $\Psi$\;
	\caption{{\substitute}($F(X,Y),\Psi,{\totalorder}$)}
	\label{reversesubstitute_algo}
\end{algorithm}
			To return the Skolem functions in terms of only $X$, {\tool} invokes {\substitute} subroutine. As described in Algorithm~\ref{reversesubstitute_algo}, for each $y_j$ of $Y$ variable, {\substitute} finds existentially quantified variables($\hat{Y}$), on which $y_j$ depends. Then, for each  $y_i$ of $\hat{Y}$; it substitutes corresponding Skolem function $\psi_i$ in the Skolem function $\psi_j$ of $y_j$.
%			For a formula $\exists Y F(X,Y)$, {\substitute} performs substitutes the Skolem function $\Psi$ as function of only $X$ variables.  For each $y_j$ of $Y$ variable, {\substitute} substitutes finds existentially quantified variables($\hat{Y}$), on which $y_j$ depends. {\substitute} sequentially substitutes $y_i$ of $\hat{Y}$ with  corresponding Skolem function $\psi_i$ in the Skolem function $\psi_j$ of $y_j$.  

\subsection{Example}
We now illustrate our algorithm though an example.

\begin{example}
	\label{example:main}
	Let $X=\{x_1,x_2\}$, $Y=\{y_1,y_2,y_3\}$ in $\exists{Y} F(X,Y)$ where $F(X,Y)$ is $(y_1 \leftrightarrow (x_1 \lor x_2)) \land (y_2 \leftrightarrow (x_1 \land (x_2 \lor y_1)))\land (y_1 \lor y_2 \lor y_3)$
\end{example}

\begin{description}
	\item [{\phaseone}:] As $y_3$ appears only as positive literal in $F$, \tool{} discovers $y_3$ as a positive unate, synthesizing a Skolem function($\psi_3$) for $y_3$ as $1$. 
	
	\item [{\phasetwo}:] 
	{\tool} generates training data through sampling (Figure \ref{tab:samples_F}) and attempts to learn a candidate Skolem function $\psi_1$ of $y_1$  by creating a decision tree (Figure \ref{fig:decision tree 1 F}). The decision tree construction uses the samples of $x_1,x_2,y_2$ as features and samples of $y_1$ as labels. This decision tree is compiled down to a propositional formula by taking a disjunction over all paths that end in leaf nodes with label $1$: in Figure \ref{fig:decision tree 1 F}, $\psi_1$ is synthesized as $(\lnot x_1 \land  x_2) \lor (x_1)$.
	
	As $\psi_1$ is not dependent on $y_2$, the samples of $\{x_1,x_2,y_1\}$ can be used to predict $y_2$. Considering the path to the leaf nodes of the tree (Figure \ref{fig:decision tree 2 F}) with label 1, $\psi_2$ is $x_1$.
	
	\item [{\phasethree}:] 
	Error formula $E(X,Y,Y')$ is {\unsat} with $\psi_1= (\lnot x_1 \land  x_2) \lor (x_1),\; \psi_2 = x_1, \; \text{and}\; \psi_3=1$. {\tool} returns $\Psi$ as skolem functions for $\exists Y F(X,Y)$.
	\par
	To illustrate {\phasethree}; let us assume that {\phasetwo} returns the candidate Skolem functions  $\psi_1 = x_1, \; \psi_2 = x_1, \; \text{and}\; \psi_3=1 $. We construct the error formula, $E(X,Y,Y') = F(X,Y) \land \lnot F(X,Y') \land (y'_{1} \leftrightarrow \psi_1) \land (y'_{2} \leftrightarrow \psi_2)  \land (y'_{3} \leftrightarrow \psi_3)$, which turns out {\sat} with counterexample $\sigma = \langle x_1 \leftrightarrow 0$, $x_2 \leftrightarrow 1$, $y_1 \leftrightarrow 1$, $y_2 \leftrightarrow 0$, $y_3 \leftrightarrow 1$, $y'_1 \leftrightarrow 0$, $y'_2 \leftrightarrow 0$, $y'_3 \leftrightarrow 1 \rangle$. 
	
	{\tool} calls {\maxsatlist} with  $F(X,Y) \land (x_1 \leftrightarrow \sigma[x_1]) \land (x_2 \leftrightarrow \sigma[x_2]) \land (y_1 \leftrightarrow \sigma[y'_1]) \land (y_2 \leftrightarrow \sigma[y'_2]) \land (y_3 \leftrightarrow \sigma[y'_3])$, which returns $Ind =\{1\}$ as $(y_1 \leftrightarrow \sigma[y'_1])$ is an unsatisfiable constraint. 
	
        Repair synthesis commences for $\psi_1$ with a query for satisfiability on $G_1= F(X,Y) \land (x_1 \leftrightarrow \sigma[x_1]) \land (x_2 \leftrightarrow \sigma[x_2]) \land (y_1 \leftrightarrow \sigma[y'_1])\land (y_2 \leftrightarrow \sigma[y'_2])$. As the formula is unsatisfiable, {\tool} calls {\findcore}, which returns variable $x_2$ (as $(x_2 \leftrightarrow \sigma[x_2])$ and  $(y_1 \leftrightarrow \sigma[y'_1])$ are not satisfiable in $G_1$).
	
	As the repaired output of $\psi_1$ on current $\sigma$ must change from 0 to 1, $\psi_1$ is updated by a disjunction of the old $\psi_1$ with $x_2$, yielding the refined formula $x_1 \lor x_2$. For this case, the formula $E(X,Y,Y') = F(X,Y) \land \lnot F(X,Y') \land (y'_{1} \leftrightarrow \psi_1) \land (y'_{2} \leftrightarrow \psi_2) \land (y'_{3} \leftrightarrow \psi_3)$ is {\unsat}, thus passing the verification check. 
\end{description}
\begin{minipage}[10cm]{\textwidth}
	\begin{minipage}[b]{0.26\textwidth}
		\centering
		\begin{tabular}{|c|c|c|c|}\hline
			$x_1$ & $x_2$ & $y_1$ & $y_2$  \\ \hline
			$0$ & $0$ & $0$ & $0$ \\ \hline
			$0$ & $1$ & $1$ & $0$  \\ \hline
			$0$ & $1$ & $1$ & $0$ \\ \hline
			$1$ & $0$ & $1$ & $1$ \\ \hline
			$1$ & $1$ & $1$ & $1$ \\ \hline
		\end{tabular}
		\captionof{figure}{\label{tab:samples_F}Samples of $F$}
	\end{minipage}
	\hfill
	\begin{minipage}[b]{0.35\textwidth}
		\centering
		\includegraphics[scale=0.28]{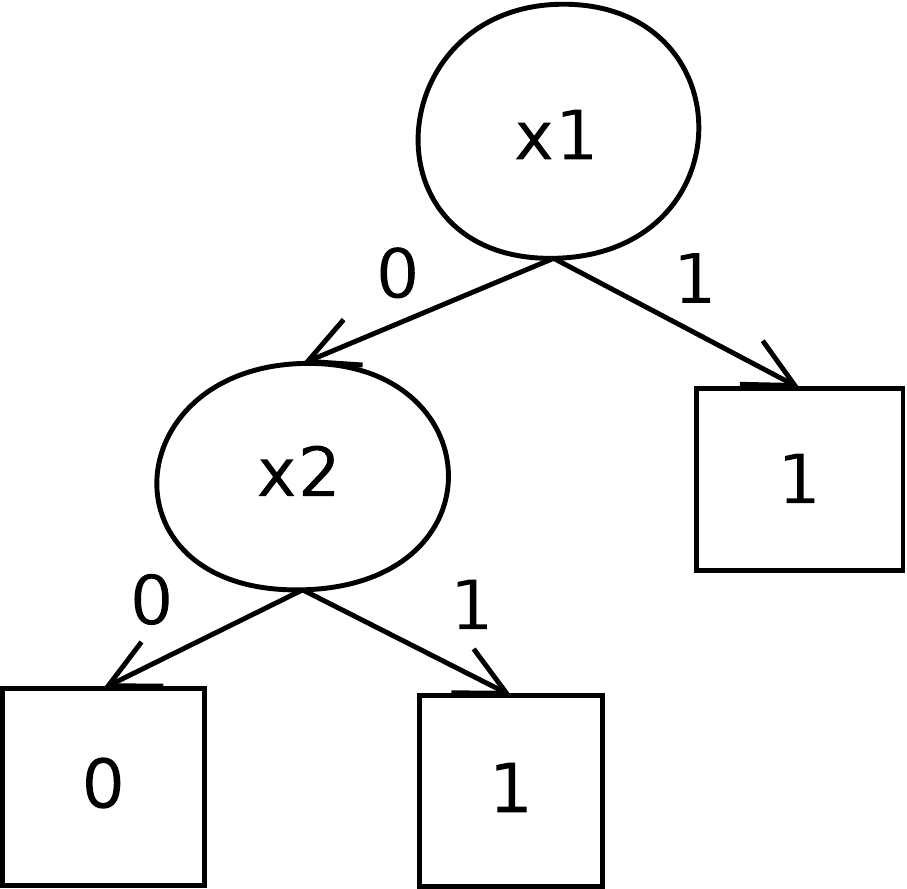}
		\captionof{figure}{\label{fig:decision tree 1 F}Decision tree for $y_1$}
	\end{minipage}
	\hfill
	\begin{minipage}[b]{0.35\textwidth}
		\centering
		\includegraphics[scale=0.28]{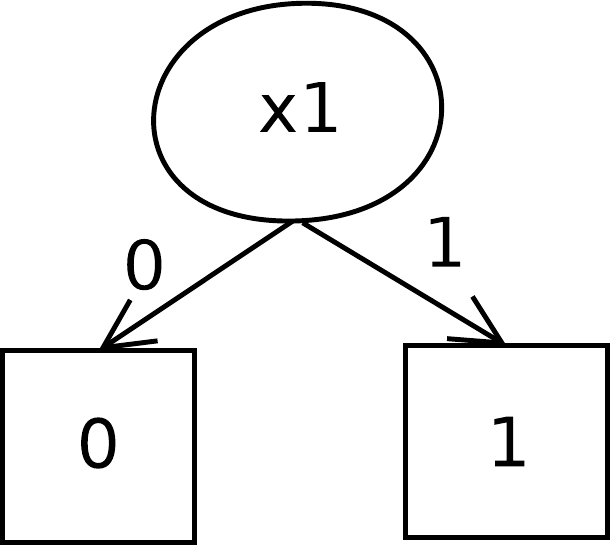}
		\captionof{figure}{\label{fig:decision tree 2 F}Decision tree for $y_2$}
	\end{minipage}
\end{minipage}

\section{Experimental Results}\label{sec:experiments}
We evaluate the performance of {\tool} on the union of all the benchmarks employed in the most recent works~\cite{AACKRS19,ACGKS18},which includes $609$ benchmarks from different sources: Prenex-2QBF track of QBFEval-17\cite{qbfeval17}, QBFEval-18\cite{qbfeval18}, disjunctive\cite{ACJS17}, arithmetic\cite{TV17} and factorization\cite{ACJS17}. We ran all the tools as per the specification laid out by their authors. We used Open-WBO~\cite{MML14} for our {\maxsat} queries and PicoSAT~\cite{B08} to compute {\unsatcore}.  We used PicoSAT for its ease of usage and we expect further performance improvements by upgrading to one of the state of the art SAT solvers. We have used the Scikit-Learn\cite{scikit-learn} to create decision trees in {\phasetwo} phase of {\tool}. We have also used ABC~\cite{abc} to represent and manipulate Boolean functions. To allow for the input formats supported by the different tools, we use the utility scripts available with the {\bfss} distribution~\cite{ACGKS18} to convert each of the instances to both QDIMACS and Verilog formats. For {\tool}, unless otherwise specified, we set the number of samples according to heuristic based on $|Y|$ as described in Section~\ref{sec:phasetwoimpact} and minimum impurity decrease to $0.005$. All our experiments were conducted on a high-performance computer cluster with each node consisting of a E5-2690 v3 CPU with $24$ cores and $96$GB of RAM, with a memory limit set to $4$GB per core. All tools were run in a single-threaded mode on a single core with a timeout of $7200$ seconds.

The objective of our experimental evaluation was two-fold: to understand the impact of various design choices on the runtime performance of {\tool} and to perform an extensive comparison of runtime performance vis-a-vis state of the art synthesis tools. In particular, we sought to answer the following questions:
	\begin{enumerate}
		\item How does the performance of {\tool} compare with state of the functional synthesis engines?
		\item How do the usage of different sampling schemes and the quality of samplers impact the performance of {\tool}?
		%\item Is the size of decision tree an important discriminator for performance of {\tool}?
		%\item How should one fix the number of samples used to generate the decision tree?
		\item What is the impact of {\phasetwo} on the performance of {\tool}?	
		\item What is the distribution of the time spent in each phase of {\tool}?
		\item How does using {\maxsat} solver to identify the potential erring Skolem functions impacts on the performance of  {\tool}?
		\item How does employing self-substitution for some Skolem functions impact {\tool}?
	\end{enumerate}

We observe that {\tool} significantly improves upon state of the art, and solves 356 benchmarks while the state of the art tool can only solve 280; in particular, {\tool} solves $60$ more benchmarks that could not be solved by any of the state of the art tools. To put the runtime performance statistics in a broader context, the number of benchmarks solved by techniques developed over the past five years range from 206 to 280, i.e., a difference of 74, which is same as an increase of 76 (i.e., from 280 to 356) due to {\tool}.
\par
Our experimental evaluation leads to interesting conclusions and several directions for future work. We observe that the performance of {\tool} is sensitive to different sampling schemes and the underlying samplers; in fact, we found that biased sampling yields better results than uniform sampling. This raises interesting questions on the possibility of designing specialized samplers for this task. Similarly, we observe interesting trade offs between the number of samples and the minimum impurity decrease in {\phasetwo}. The diversity of our extensive benchmark suite produces a nuanced picture with respect to time distribution across different phases, highlighting the critical nature of each of the phases to the performance of {\tool}. {\tool} shows significant performance improvement by using {\maxsat} solver to identify candidates to refine. {\tool} also has significant performance improvement with self substitution in terms of the required number of refinements.   %(we defer detailed discussion on this to the Appendix in supplementary material).

\subsection{Comparison with other tools} 
\begin{table}
\caption{\label{tab:benchmarks_solved} No. of benchmarks solved by different tools}
\centering
\begin{tabular}{ccccccccc}\toprule
 Total & {\bafsyn} & {\caqe} & {\depqbf} & {\ctosyn} & {\bfss} & {\cadet} & \textbf{{\tool}} & All Tools \\ \midrule
 $609$ & $13$ & $54$ & $59$ &$206$ & $247$ & $280$ & $\boldsymbol{356}$  & $476$\\ \bottomrule
\end{tabular}
\end{table}

\begin{figure}[t]
	\begin{center}
		\includegraphics[scale=0.49]{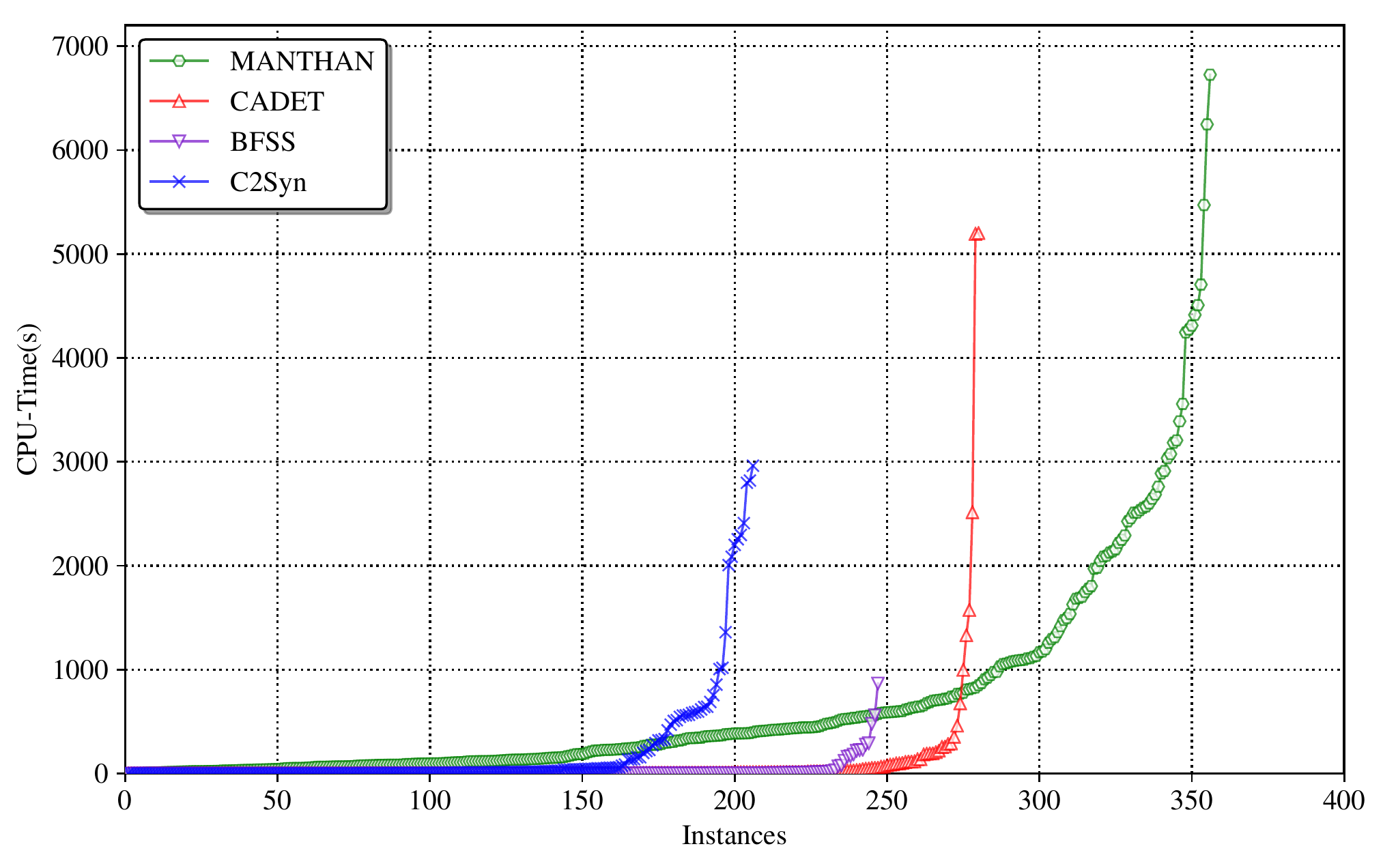}
		\caption{\label{fig:manthanvsothertool}{\tool} versus competing tools for Skolem function synthesis}
	\end{center}
\end{figure}
 We now present performance comparison of {\tool} with the current state of the art synthesis tools, {\bfss}~\cite{ACGKS18}, {\ctosyn}~\cite{AACKRS19}, {\bafsyn}~\cite{CFTV18} and the current state of the art 2-QBF solvers {\cadet}~\cite{R19},{\caqe}~\cite{RT15} and {\depqbf}~\cite{LE17}. The certifying 2-QBF solver produces QBF certificates, that can be used to extract Skolem functions~\cite{BJ11}. Developers of {\bafsyn} and {\depqbf} confirmed that the tools produce Skolem function for only valid instances, i.e. when $\forall X \exists Y F(X,Y)$ is valid. Note that the current version of {\caqe} does not support certification and we have used {\caqe} version 2 for the experiments after consultation with the developers of {\caqe}.
 We present the number of instances solved Table~\ref{tab:benchmarks_solved}. Out of 609 benchmarks, the most number of instances solved by any of the remaining techniques is $280$ while {\tool} is able to solve $356$ instances -- a significant improvement over state of the art. We will focus on top $4$ synthesis tools from Table~\ref{tab:benchmarks_solved} for further analysis. 
   \par
For a deeper analysis of runtime behavior, we present the cactus plot in Figure~\ref{fig:manthanvsothertool}: the number of instances are shown on the $x$-axis and the time taken on the $y$-axis; a point $(x, y)$ implies that a solver took less than or equal to $y$ seconds to find Skolem function of $x$ instances on a total of $609$ instances. An interesting behavior predicted by cactus plot and verified upon closer analysis is that for instances that can be solved by most of the tools, the initial overhead due to a multi-phase approach may lead to relatively larger runtime for {\tool}. However, with the rise in empirically observed hardness of instances, one can observe the strengths of the multi-phase approach. Overall, {\tool} solves $76$ more instances than the rest of the remaining techniques. 
\begin{table}
	\centering
	\caption[]{\label{tab:takneek_comparison}Manthan vs other state-of-the-art tools}
	\begin{tabular}{cccccccccc}%\cline{3-6} 
		\toprule
		
		\multicolumn{2}{c} {} & {\ctosyn} & {\bfss} & {\cadet} & All Tools \\ \midrule
		%\cline{1-6}
		\multirow{2}{*}{{\tool}} & Less & $13$ & $85$ & $111$ & $122$ \\ 
		%\cline{2-6}
		& More & $\boldsymbol{163}$ & $\boldsymbol{194}$ & $\boldsymbol{187}$ & $\boldsymbol{60}$ \\ \bottomrule
		
	\end{tabular}
	
\end{table}

%   

%	For ``easier" benchmarks, {\tool} takes more time as compared to the competing tools as it expends time in gathering data (sampling) and generating candidate skolem function. However, overall, it manages to solve  many more($109$) benchmarks than any state-of-art-tool, showing the value of data-driven formal method algorithms.
	%({\cadet}, {\bfss} and {\ctosyn} solve $191$, $247$, $206$ benchmarks respectively). 
	We show a pairwise comparison of {\tool} vis-a-vis other techniques in Table~\ref{tab:takneek_comparison}. The second row of the table lists the number of instances that were solved by the technique in the corresponding column but not by {\tool} while the third row lists the number of instances that were solved by {\tool} but not the corresponding technique. First, we observe that {\tool} solves $163$, $194$, and $187$ instances that are not solved by {\ctosyn}, {\bfss}, and {\cadet} respectively.
	Though {\bfss} and {\cadet} solve more than 80 instances that {\tool} does not solve, they are not complementary; there are only $121$ instances that can be solved by either {\bfss} or {\cadet} but {\tool} fails to solve. A closer analysis of {\tool}'s performance on these instances revealed that the decision trees generated by {\candidateskf} were shallow, which is usually a sign of significant under-fitting. On the other hand, there are $130$ instances that {\tool} solves, but neither {\cadet} nor {\bfss} can solve. These instances have high dependencies between variables that {\tool} can infer from the samples en route to predicting good candidate Skolem functions. 
	Akshay et al. \cite{AACKRS19} suggest that {\ctosyn} is an orthogonal approach to {\bfss}. {\tool} solves $81$ instances that neither {\ctosyn} nor {\bfss} is able to solve, and these tools together solve $86$ instances that {\tool} fails to solve.
	Overall, {\tool} solves $\boldsymbol{60}$ instances beyond the reach of any of the above state of the art tools. 
	
\subsection{Impact of the sampling scheme}

	To analyze the impact of the adaptive sampling and the quality of distributions generated by underlying samplers, we augmented {\tool} with samples drawn from different samplers for adaptive and non-adaptive sampling. In particular, we employed QuickSampler~\cite{DLBS18}, KUS~\cite{SGRM18}, UniGen2~\cite{CMV14},  and {\sampler}\footnote{{\sampler} is developed by Mate Soos and Kuldeep S. Meel, and is pending publication.}. The samplers KUS and UniGen2 could only produce samples for mere $14$ and $49$ benchmarks respectively within a timeout of $3600$ seconds.  Hence, we have omitted KUS and UniGen2 from further analysis. We also experimented with a naive enumeration of solution using off-the-shelf SAT solver, CryptoMiniSat~\cite{cryptominisat}. It is worth noting that QuickSampler performs worse than {\sampler} for uniformity testing using Barbarik~\cite{CM19}. In our implementation, we had to turn off the validation phase of QuickSampler to allow generation number of samples within a reasonable time.
	To statistically validate our intuition described in Section~\ref{sec:overview}, we performed  adaptive sampling using {\sampler}. We use {\adasampler} to refer to the adaptive sampling implementation. 
%	We also designed a new sampling scheme wherein we assigned different weights for the existentially and universally quantified positive literals. We used an in-house (currently unpublished) weighted sampler, {\sampler}, in two modes:
%	\begin{itemize}
%		\item {\bf {\sampler} with Bias($0.5,0.5$)} [unbiased mode] we assigned a uniform weight of {$0.5$} to all the universally quantified positive literals and as well as the existentially quantified positive literals;
%		\item {\bf {\sampler} with Bias($0.5,q$)} [biased mode] to perform a biased sampling, setting weights $0.5$ to each of universally quantified positive literals and $q$ is an adaptive bias(section ~\ref{overview}) for the existentially quantified positive literals. 
%	\end{itemize}
%	
	
\begin{figure}
			\begin{minipage}[t]{\textwidth}
				\begin{minipage}[t]{0.35\textwidth}
					\vspace{-2pt}
					\small
					\includegraphics[scale=0.3]{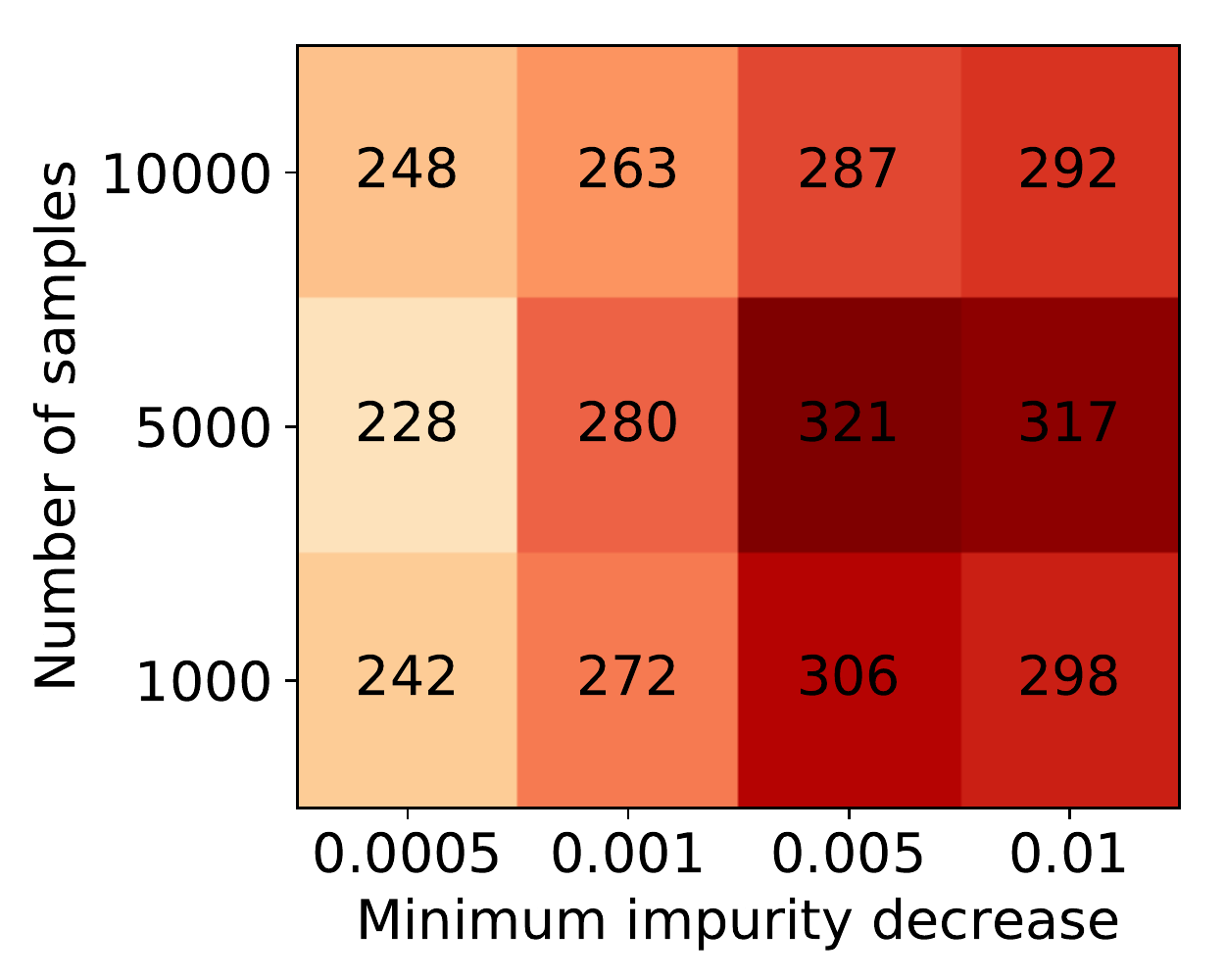}
					\captionof{figure}{\label{fig:manthandiffheu}Heatmap of \# instances solved.(Best viewed in color)}
				\end{minipage}
				%	\begin{table}
				\begin{minipage}[t]{0.6\textwidth}
					\vspace{0pt}
					\footnotesize
						\begin{tabular}{ccccc}\toprule
							\multirow{2}{2em}{Sampler} & \multicolumn{3}{c}{No. of instances solved} & \multirow{2}{4em}{\#Solved}   \\
							\cline{2-4} 
							& {\phaseone} & {\phasetwo} & {\phasethree} & 
							\\\midrule
							CryptoMiniSat& $66$ & $14$ & $191$ & $271$ \\ 
							QuickSampler & $66$ & $28$ & $181$ & $275$ \\
							{\sampler} & $66$ & $51$ & $228$ & $345$ \\ 
							{\adasampler} & $66$ & $66$ & $224$ & $\boldsymbol{356}$ \\ \bottomrule
						\end{tabular}
					\captionof{table}{\tool{} with different samplers}\label{tab:takneeksample_comparison}
				\end{minipage}
				%	\end{table}
				%		\hfill
				%		\begin{minipage}[b]{0.55\textwidth}
				%			\small
				%			\centering
				%			\includegraphics[width=\textwidth,height=4cm]{figures/manthan_meantime.pdf}
				%			\captionof{figure}{\label{fig:manthantimeoverbenchmarks} {\tool} total time divisions for different benchmarks sets.(Best viewed in color)}
				%		\end{minipage}
				%	\end{figure}
			\end{minipage}	
		\end{figure}

	Table~\ref{tab:takneeksample_comparison} presents the performance of {\tool} with different samplers listed in Column 1. The columns 2, 3, and 4 lists the number of instances that were solved during the execution of respective phases: {\phaseone}, {\phasetwo}, and {\phasethree}. Finally, column 5 lists the total number of instances solved. Two important findings emerge from Table~\ref{tab:takneeksample_comparison}: Firstly, as the quality of samplers improve, so does the performance of {\tool}. In particular, we observe that with the improvement in the quality of samples leads to {\tool} solving more instances in {\phasetwo}. Secondly, we see a significant increase in the number of instances that can be solved due to {\phasetwo} with samples from {\adasampler}. It is worth remarking that one should view the adaptive scheme proposed in Section~\ref{sec:overview} to be a proof of concept and our results will encourage the development of more complex schemes.  \newline
	
%	Our new sampling scheme, {\sampler} with Bias($0.5,q$), is able to solve the maximum number of instances within the provided timeout of $7200$ seconds showing the utility of biasing the existential variables towards 1 (section ~\ref{overview}). {\sampler} with Bias($0.5, 0.5$) also performs well due to its ability to produce \textit{nearly} uniform samples. The samples from QuickSampler are not only far from uniform~\cite{CM19}, . Incremental sampling with CrypoMiniSat performs the worst; with no effort at uniform sampling, a {\sat} solver is likely to fail to cover the behaviors of the function well.
	\subsection{Impact of {\phasetwo}}\label{sec:phasetwoimpact}
	
	To analyze the impact of different design choices in {\phasetwo}, we analyzed the performance of {\tool} for different samples ($1000$, $5000$ and $10000$) generated by {\getsamples} and for different choices of minimum impurity decrease ($0.001$, $0.005$, $0.0005$). Figure~\ref{fig:manthandiffheu} shows a heatmap on the number of instances solved on each combination of the hyperparameters; the closer the color of a cell is  to the red end of the spectrum, the better the performance of {\tool}.

	At the first look, Figure~\ref{fig:manthandiffheu} presents a puzzling picture: It seems that increasing the number of samples does not improve the performance of {\tool}. On a closer analysis, we found that the increase in the number of samples leads to an increase in the runtime of {\candidateskf} but without significantly increasing the number of instances solved during {\phasetwo}. The runtime of {\candidateskf} is dependent on the number of samples and $|Y|$. On the other hand, we see an interesting trend with respect to minimum impurity decrease where the performance first improves and then degrades. A plausible explanation for such a behavior is that with an increase in {\em minimum impurity decrease}, the generated decision trees tend to underfit while significantly low values of {\em minimum impurity decrease} lead to overfitting. We intend to study this in detail in the future.
	
	%KSM: Is the reason for minimum impurity also same as number of samples
	
	Based on the above observations, we set the value of minimum impurity decrease to $0.005$ and set the number of samples to (1) $10000$ for $|Y| < 1200$, (2) 5000 for $1200 < |Y| \leq 4000$, and (3) 1000 for  $|Y| > 4000$.

\subsection{Division of time taken across different phases}
    \begin{figure}[t]
	\small
	\centering
	\includegraphics[scale=0.40]{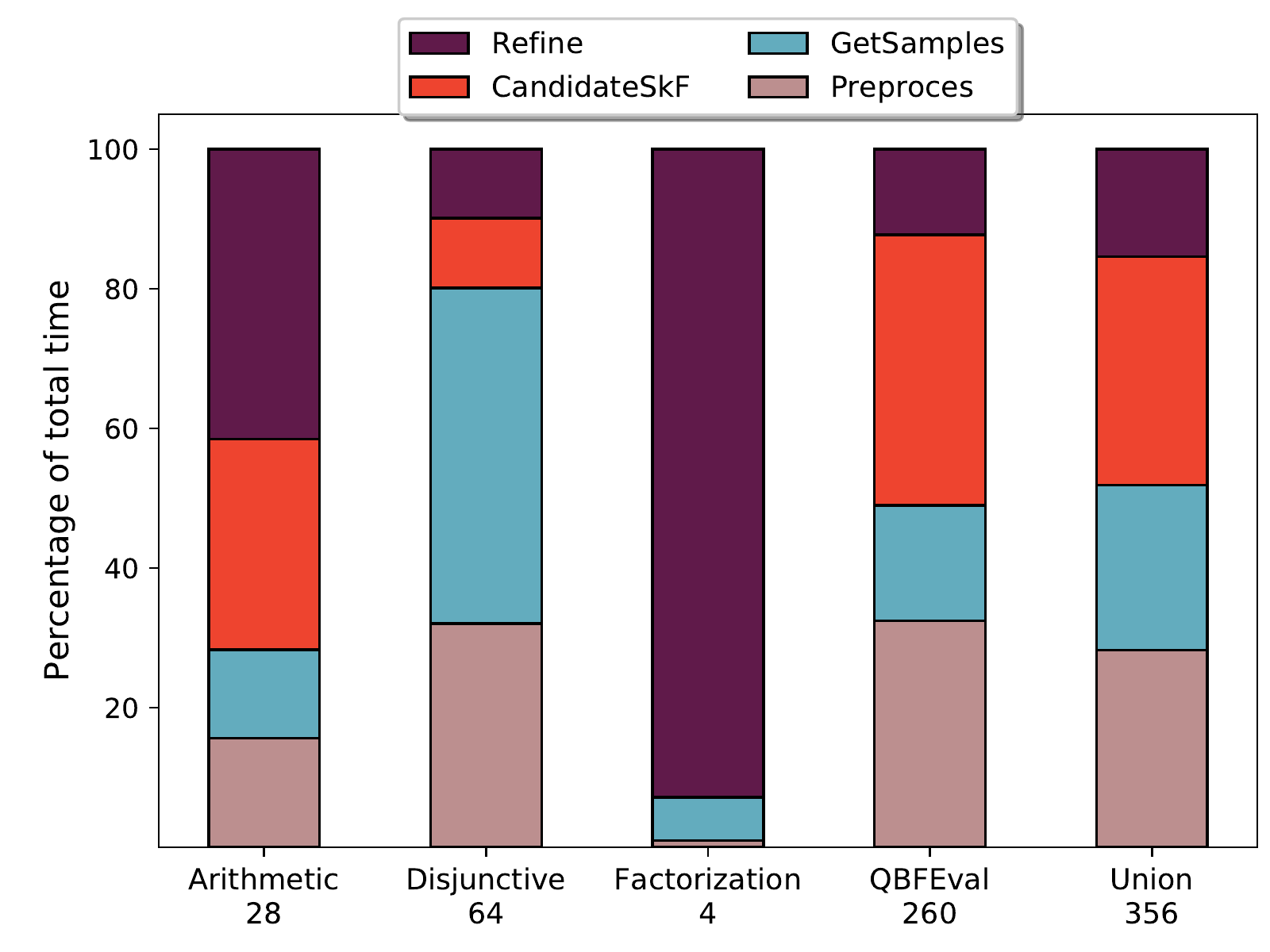}
    \captionof{figure}{\label{fig:manthantimeoverbenchmarks} Fraction of time spent in different phases in {\tool} over different classes of benchmarks. (Best viewed in color)}
\end{figure}
To analyze the time taken by different phases of {\tool} across different categories of the benchmarks, we normalize the time taken for each of the four core subroutines, {\preprocess}, {\getsamples}, {\candidateskf}, and {\refine}, for every benchmark that was solved by {\tool} such that the sum of time taken for each benchmark is 1. We then compute the mean of the normalized times across different categories instances. Figure~\ref{fig:manthantimeoverbenchmarks} shows the distribution of mean normalized times for different categories: Arithmetic, Disjunction, Factorization, QBFEval, and all the instances.  
%Figure \ref{fig:manthantimeoverbenchmarks} shows the fraction of time taken by the different subroutines of {\tool} for different categories of solved benchmarks. We used the median of time taken by each subroutine over a benchmark set to get a relative percentage of time.

The diversity of our benchmark suite shows a nuanced picture and shows that the time taken by different phases strongly depends on the family of instances. For example, the disjunctive instances are particularly hard to sample and an improvement in the sampling techniques would lead to significant performance gains. On the other hand, a significant fraction of runtime is spent in the {\candidateskf} subroutine indicating the potential gains due to improvement in decision tree generation routines. In all, Figure~\ref{fig:manthantimeoverbenchmarks} identifies the categories of instances that would benefit from algorithmic and engineering improvements in {\tool}'s different subroutines. 
\subsection{Impact of using {\maxsat}}
In {\refine}, {\tool} invokes the {\maxsatlist} subroutine, which calls {\maxsat} solver to identify the potential erring Skolem functions. To observe the impact of using {\maxsat} solver to identify the candidates to refine, we did an experiment with {\tool}, without {\maxsatlist} subroutine call. For all $y_i$, where $\sigma[y_i] \neq \sigma[y'_i]$ were considered as candidates to refine. {\tool} without {\maxsatlist} subroutine call solved $204$ instances that represents a significant drop in  the number of solved instances by {\tool} with {\maxsatlist} subroutine.
	\subsection{Impact of self-substitution}
	\begin{figure}
		\centering
		\begin{minipage}{\textwidth}
		\begin{minipage}{.48\textwidth}
				\subfigure[Benchmark \textit{pdtpmsmiim-all-bit:} plot for no. of Skolem functions vs required no. of refinements\label{fig:self-sub-a}]{\includegraphics[width=\textwidth]{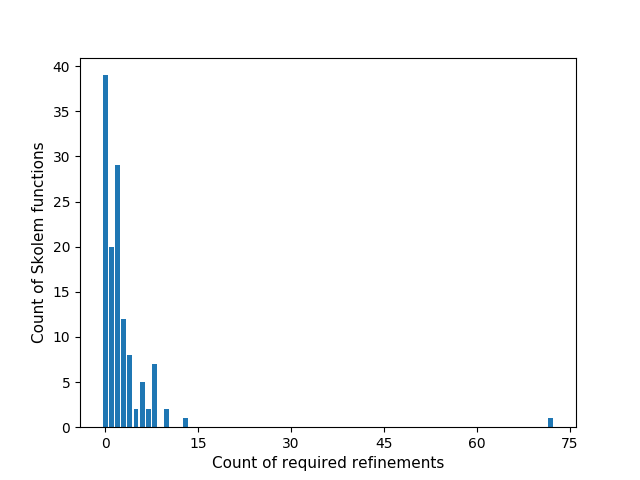}}
		\end{minipage}
		\hspace{1mm}
		\begin{minipage}{.48\textwidth}
			\subfigure[Benchmark \textit{pdtpmsmiim:} plot for no. of Skolem functions vs required no. of refinements\label{fig:self-sub-b}]{\includegraphics[width=\textwidth]{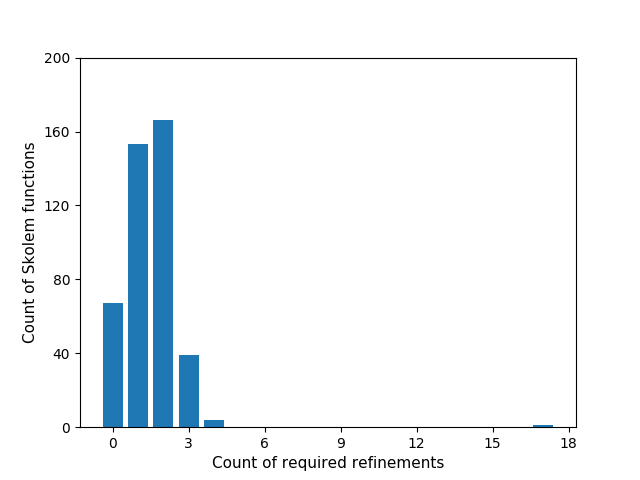}}
		\end{minipage}
		\end{minipage}
	
%		\begin{subfigure}{.48\textwidth}
%			\centering\includegraphics[scale=0.38]{figures/selfsub1.png}
%			\caption{\label{fig:self-sub-a}Benchmark \textit{pdtpmsmiim-all-bit:} plot for no. of Skolem functions vs required no. of refinements}
%		\end{subfigure}
%		\begin{subfigure}{.48\textwidth}
%			\centering\includegraphics[scale=0.38]{figures/selfsub2.png}
%			\caption{\label{fig:self-sub-b}Benchmark \textit{pdtpmsmiim:} plot for no. of Skolem functions vs required no. of refinements}
%		\end{subfigure}
		\caption{\label{fig:self-sub} The plots to show the required number of refinements for the candidate Skolem functions.}
	\end{figure}
    To understand the impact of self-substitution, we profile the behavior of candidate Skolem functions with respect to number of refinements for two of our benchmarks; \textit{pdtpmsmiim-all-bit} and \textit{pdtpmsmiim}. In Figure~\ref{fig:self-sub}, we use histograms with the number of candidate Skolem functions on y-axis and required number of refinements on x-axis. A bar of height $a$ i.e $y=a$ at $b$ i.e $x=b$ in Figure~\ref{fig:self-sub} represents that $a$ candidate Skolem functions converged in $b$ refinements. The histograms show that only a few Skolem functions require a large number of refinements: the tiny bar towards the right end in Figure~\ref{fig:self-sub-a} represents that for the benchmark \textit{pdtpmsmiim-all-bit} only $1$ candidate Skolem function required more than $60$ refinements whereas all other candidate Skolem functions needed less than $15$ refinements. Similarly, for the benchmark \textit{pdtpmsmiim}, Figure~\ref{fig:self-sub-b} shows that only $1$ candidate Skolem function was refined more than $15$ times, whereas all other Skolem functions required less than $5$ refinements. We found similar behaviors in many of our other benchmarks. 
	 \par 
    Based on the above trend and an examination of the decision trees corresponding to these instances, we hypothesize that some Skolem functions are hard to learn through data. For such functions, the candidate Skolem function generated from the data-driven phase in {\tool} tends to be poor, and hence {\tool} requires a long series of refinements for convergence. Since our refinement algorithm is designed for small, efficient corrections, we handle such hard to learn Skolem functions by synthesizing via self-substitution. {\tool} detects such functions via a threshold on the number of refinements, which is empirically determined as 10, to identify  hard to learn instances and sets them up for self-substitution.
	\par
%	We observed that the primary reason behind repeated refinements was the lack of good guesses of candidate Skolem functions. The decision trees for those existentially quantified variables were not able to learn candidate Skolem functions as they had just one node. {\tool} finds such candidate Skolem functions in {\phasetwo} phase and puts corresponding existentially quantified variables at the last in the dynamic variable ordering. {\tool} preforms self-substitution for those existentially quantified variable, if  corresponding candidate Skolem function requires refinement more than a threshold, which is heuristically set to $10$ in our experiment. 
	In our experiments, we found $75$ instances out of $356$ solved instances required self-substitution, and for $51$ of these $75$ instances, only one variable undergoes self-substitution. Table~\ref{tab:selfsub} shows the impact of self-substitution for five of our benchmarks: {\tool} has significant performance improvement with self-substitution in terms of the required number of refinements, which in turns affects the overall time. Note that {\tool} can refine multiple candidates in a single {\refine} call. For the first four benchmarks, all the other Skolem function except the poor candidates were synthesized earlier than $10$ refinement iteration, and at the $10^{th}$ refinement iteration the poor candidate functions hit our threshold for self-substitution.
	Taking the case of the last benchmark, all the other Skolem functions for it were synthesized earlier than 40 refinement cycles, and the last $16$ iterations were only needed for $2$ of the poor candidate functions to hit our threshold for self-substitution. Note that self-substitution can lead to an exponential blowup in the size of the formula, but it works quite well in our design as most Skolem functions are learnt quite well in the {\phasetwo} phase.
\begin{table}[t]
	\centering
	\caption[]{\label{tab:selfsub}{\tool} : Impact of self substitution}
	\begin{tabular}{cccccccc}\toprule
		\multirow{3}{5em}{Benchmarks $\exists Y F(X,Y)$} & \multirow{3}{2em}{$|X|$} & \multirow{3}{2em}{$|Y|$} & \multicolumn{2}{c}{No. of Refinements} & &\multicolumn{2}{c}{Time(s)} \\
	
		\cline{4-5} \cline{6-8}
		& & & \multicolumn{2}{c}{Self-Substitution} & & \multicolumn{2}{c}{Self-Substitution} \\
		& &	& Without  & With& & Without & With  
		\\ \midrule
		kenflashpo2-all-bit & $71$ & $32$ & $319$ & $10$ & & $35.88$ & $19.22$ \\
		eijkbs1512 & $316$ & $29$ & $264$ & $10$ & & $42.88$  & $32.35$ \\
		pdtpmsmiim-all-bit & $429$ & $30$ & $313$ & $10$ & & $72.75$ & $36.08$ \\ 
		pdtpmssfeistel & $1510$  & $68$ & $741$ & $10$ & & $184.11$ & $115.07$ \\
		pdtpmsmiim & $418$ & $337$ & $127$ & $56$ & & $1049.29$ & $711.48$ \\
		\bottomrule
	\end{tabular}
	
\end{table}

\section{Conclusion}
\label{sec:conclusion}
Boolean functional synthesis is a fundamental problem in Computer Science with a wide variety of applications. In this work, we propose a novel data-driven approach to synthesis that employs constrained sampling techniques for generation of data, machine learning for candidate Skolem functions, and automated reasoning to verify and refine to generate Skolem functions. Our approach achieves significant performance improvements. As pointed out in Section~\ref{sec:algorithm} and~\ref{sec:experiments}, our work opens up several interesting directions for future work at the intersection of machine learning, constrained sampling, and automated reasoning.
\paragraph{Acknowledgment}
We are grateful to the anonymous reviewers for constructive
comments that significantly improved the final version of the paper. We are grateful to Mate Soos for tweaking {\sampler} to support {\tool}. We are indebted to S. Akshay, Supratik Chakraborty, and Shetal Shah for their patient responses to our tens of queries regarding prior work.  We are grateful to Dror Fried and the anonymous reviewers for constructive criticism of the early draft of the work. 

This work was supported in part by National Research Foundation Singapore under its NRF Fellowship Programme[NRF-NRFFAI1-2019-0004 ] and AI Singapore Programme [AISG-RP-2018-005], and NUS ODPRT Grant [R-252-000-685-13]. The computational work for this article was performed on resources of the National Supercomputing Centre, Singapore: \url{https://www.nscc.sg}~\cite{nscc}.

%\subh{take care of capitalizations; make sure titles are correct ([1] and [2] read the same); check the names of the conferences ([34] is ESEC/FSE not FSC)}
\clearpage
\bibliographystyle{splncs04}
\bibliography{ref}

\begin{thebibliography}{10}
\providecommand{\url}[1]{\texttt{#1}}
\providecommand{\urlprefix}{URL }
\providecommand{\doi}[1]{https://doi.org/#1}

\bibitem{nscc}
{ASTAR, NTU, NUS, SUTD}: {N}ational {S}upercomputing {C}entre {(NSCC)}
  {S}ingapore (2018), \url{https://www.nscc.sg/about-nscc/overview/}

\bibitem{qbfeval17}
{QBF} solver evaluation portal 2017, \url{http://www.qbflib.org/qbfeval17.php}

\bibitem{qbfeval18}
{QBF} solver evaluation portal 2018, \url{http://www.qbflib.org/qbfeval18.php}

\bibitem{AACKRS19}
Akshay, S., Arora, J., Chakraborty, S., Krishna, S., Raghunathan, D., Shah, S.:
  Knowledge compilation for boolean functional synthesis. In: Proc. of FMCAD
  (2019)

\bibitem{ACGKS18}
Akshay, S., Chakraborty, S., Goel, S., Kulal, S., Shah, S.: What’s hard about
  boolean functional synthesis? In: Proc. of CAV (2018)

\bibitem{ACJS17}
Akshay, S., Chakraborty, S., John, A.K., Shah, S.: Towards parallel boolean
  functional synthesis. In: Proc. of TACAS (2017)

\bibitem{ABJMRSSSTU13}
Alur, R., Bodik, R., Juniwal, G., Martin, M.M., Raghothaman, M., Seshia, S.A.,
  Singh, R., Solar-Lezama, A., Torlak, E., Udupa, A.: Syntax-guided synthesis.
  In: Proc. of FMCAD (2013)

\bibitem{BJ11}
Balabanov, V., Jiang, J.H.R.: Resolution proofs and skolem functions in {QBF}
  evaluation and applications. In: Proc. of CAV (2011)

\bibitem{BJ12}
Balabanov, V., Jiang, J.H.R.: Unified {QBF} certification and its applications.
  In: Proc. of FMCAD (2012)

\bibitem{BPR16}
Bavishi, R., Pandey, A., Roy, S.: {To} be precise: regression aware debugging.
  In: Proc. of OOPSLA (2016)

\bibitem{B08}
Biere, A.: {PicoSAT} essentials. Proc. of JSAT  (2008)

\bibitem{boole1847}
Boole, G.: The mathematical analysis of logic. Philosophical Library (1847)

\bibitem{CM19}
Chakraborty, S., Meel, K.S.: On testing of uniform samplers. In: Proc. of AAAI
  (2019)

\bibitem{CFTV18}
Chakraborty, S., Fried, D., Tabajara, L.M., Vardi, M.Y.: Functional synthesis
  via input-output separation. In: Proc. of FMCAD (2018)

\bibitem{CMV14}
Chakraborty, S., Meel, K.S., Vardi, M.Y.: Balancing scalability and uniformity
  in {SAT} witness generator. In: Proc. of DAC (2014)

\bibitem{DLBS18}
Dutra, R., Laeufer, K., Bachrach, J., Sen, K.: Efficient sampling of {SAT}
  solutions for testing. In: Proc. of ICSE (2018)

\bibitem{ENDGM18}
Ezudheen, P., Neider, D., D'Souza, D., Garg, P., Madhusudan, P.: Horn-{ICE}
  learning for synthesizing invariants and contracts. In: Proc. of OOPSLA
  (2018)

\bibitem{FG19}
Fedyukovich, G., Gupta, A.: Functional synthesis with examples. In: Proc. of CP
  (2019)

\bibitem{FTV16}
Fried, D., Tabajara, L.M., Vardi, M.Y.: {BDD}-based boolean functional
  synthesis. In: Proc. of CAV (2016)

\bibitem{GLMN14}
Garg, P., L{\"o}ding, C., Madhusudan, P., Neider, D.: {ICE}: A robust framework
  for learning invariants. In: Proc. of CAV (2014)

\bibitem{GNMR16}
Garg, P., Neider, D., Madhusudan, P., Roth, D.: Learning invariants using
  decision trees and implication counterexamples. In: Proc. of POPL (2016)

\bibitem{GLST05}
Grumberg, O., Lerda, F., Strichman, O., Theobald, M.: Proof-guided
  underapproximation-widening for multi-process systems. In: Proc. of POPL
  (2005)

\bibitem{GSRM19}
Gupta, R., Sharma, S., Roy, S., Meel, K.S.: {WAPS}: Weighted and projected
  sampling. In: Proc. of TACAS (2019)

\bibitem{HSB14}
Heule, M.J., Seidl, M., Biere, A.: Efficient extraction of skolem functions
  from {QRAT} proofs. In: Proc. of FMCAD (2014)

\bibitem{J18}
Janota, M.: Towards generalization in {QBF} solving via machine learning. In:
  Proc. of AAAI (2018)

\bibitem{JMF14}
Jo, S., Matsumoto, T., Fujita, M.: {SAT}-based automatic rectification and
  debugging of combinational circuits with lut insertions. Proc. of IPSJ T-SLDM
   (2014)

\bibitem{JSCTA15}
John, A.K., Shah, S., Chakraborty, S., Trivedi, A., Akshay, S.: Skolem
  functions for factored formulas. In: Proc. of FMCAD (2015)

\bibitem{JM11}
Jose, M., Majumdar, R.: Cause clue clauses: error localization using maximum
  satisfiability. In: Proc. of PLDI (2011)

\bibitem{JBSKW07}
Jussila, T., Biere, A., Sinz, C., Kr{\"o}ning, D., Wintersteiger, C.M.: A first
  step towards a unified proof checker for {QBF}. In: Proc. of SAT (2007)

\bibitem{abc}
Logic, B., Group, V.: {ABC}: A system for sequential synthesis and
  verification, \url{http://www.eecs.berkeley.edu/~alanmi/abc/}

\bibitem{LE17}
Lonsing, F., Egly, U.: Depqbf 6.0: A search-based {QBF} solver beyond
  traditional {QCDCL}. In: Proc. of CADE (2017)

\bibitem{L1910}
L\"{o}wenheim, L.: \"{U}ber die aufl{\"o}sung von gleichungen im logischen
  gebietekalkul. Mathematische Annalen  (1910)

\bibitem{MML14}
Martins, R., Manquinho, V., Lynce, I.: {Open-WBO}: A modular {MaxSAT} solver.
  In: Proc. of SAT (2014)

\bibitem{MM20}
Massacci, F., Marraro, L.: Logical cryptanalysis as a {SAT} problem. Journal of
  Automated Reasoning  (2000)

\bibitem{NPLSB12}
Niemetz, A., Preiner, M., Lonsing, F., Seidl, M., Biere, A.: Resolution-based
  certificate extraction for {QBF}. In: Proc. of SAT (2012)

\bibitem{scikit-learn}
Pedregosa, F., Varoquaux, G., Gramfort, A., Michel, V., Thirion, B., Grisel,
  O., Blondel, M., Prettenhofer, P., Weiss, R., Dubourg, V., Vanderplas, J.,
  Passos, A., Cournapeau, D., Brucher, M., Perrot, M., Duchesnay, E.:
  {S}cikit-learn: {M}achine {L}earning in {P}ython. Proc. of Machine Learning
  Research  (2011)

\bibitem{Q86}
Quinlan, J.R.: Induction of decision trees. Proc. of Machine learning  (1986)

\bibitem{R19}
Rabe, M.N.: Incremental determinization for quantifier elimination and
  functional synthesis. In: Proc. of CAV (2019)

\bibitem{RT15}
Rabe, M.N., Tentrup, L.: {CAQE}: A certifying {QBF} solver. In: Proc. of FMCAD
  (2015)

\bibitem{RTRS18}
Rabe, M.N., Tentrup, L., Rasmussen, C., Seshia, S.A.: Understanding and
  extending incremental determinization for {2QBF}. In: Proc. of CAV (2018)

\bibitem{SGRM18}
Sharma, S., Gupta, R., Roy, S., Meel, K.S.: Knowledge compilation meets uniform
  sampling. In: Proc. of LPAR (2018)

\bibitem{cryptominisat}
Soos, M.: msoos/cryptominisat (2019),
  \url{https://github.com/msoos/cryptominisat}

\bibitem{SGF13}
Srivastava, S., Gulwani, S., Foster, J.S.: Template-based program verification
  and program synthesis. STTT  (2013)

\bibitem{TV17}
Tabajara, L.M., Vardi, M.Y.: Factored boolean functional synthesis. In: Proc.
  of FMCAD (2017)

\bibitem{VR17}
Verma, S., Roy, S.: Synergistic debug-repair of heap manipulations. In: Proc.
  of ESEC/FSE (2017)

\end{thebibliography}
%
%\clearpage
%\appendix
%\input{chapters/appendix}

\end{document}